\documentclass{article}
\pdfoutput=1
\PassOptionsToPackage{numbers, compress}{natbib}

\usepackage[final]{neurips_2022}

\usepackage[utf8]{inputenc} 
\usepackage[T1]{fontenc}    
\usepackage{url}            
\usepackage{booktabs}       
\usepackage{amsfonts}       
\usepackage{nicefrac}       
\usepackage{microtype}      
\usepackage{xcolor}         

\usepackage{array}
\usepackage{multirow}
\usepackage{microtype}
\usepackage{wrapfig}

\usepackage[unicode=true,bookmarks=false,breaklinks=false,pdfborder={0 0 1},backref=section,colorlinks=false]{hyperref}
\usepackage{enumitem}

\usepackage{graphicx}
\usepackage{pifont}
\usepackage{colortbl}
\definecolor{mygray}{gray}{.9}
\definecolor{mypink}{rgb}{.99,.91,.95}
\definecolor{mycyan}{cmyk}{.1,0,0,0}
\newcommand{\cmark}{\ding{51}}%
\newcommand{\xmark}{\ding{55}}%
\newcommand{\xmarkg}{\textcolor{lightgray}{\ding{55}}}%

\title{DetCLIP: Dictionary-Enriched Visual-Concept

Paralleled Pre-training for Open-world Detection}

\author{%
  Lewei Yao$^{1,2}$$^*$, Jianhua Han$^2$$^*$, Youpeng Wen$^3$, Xiaodan Liang$^3$, Dan Xu$^1$, \\\textbf{Wei Zhang$^2$, Zhenguo Li$^2$, Chunjing Xu$^2$, Hang Xu$^2$$^\dagger$} \\
  $^1$Hong Kong University of Science and Technology, $^2$Huawei Noah's Ark Lab \\
  $^3$Shenzhen Campus of Sun Yat-Sen University
}

\begin{document}

\maketitle

\begin{abstract}
{Open-world object detection, as a more general and challenging goal, aims to recognize and localize objects described by arbitrary category names. The recent work GLIP formulates this problem as a grounding problem by concatenating all category names of detection datasets into sentences, which leads to inefficient interaction between category names. This paper presents DetCLIP, a paralleled visual-concept pre-training method for open-world detection by resorting to knowledge enrichment from a designed concept dictionary. To achieve better learning efficiency, we propose a novel paralleled concept formulation that extracts concepts separately to better utilize heterogeneous datasets (i.e., detection, grounding, and image-text pairs) for training. We further design a concept dictionary~(with descriptions) from various online sources and detection datasets to provide prior knowledge for each concept. By enriching the concepts with their descriptions,
we explicitly build the relationships among various concepts to facilitate the open-domain learning. The proposed concept dictionary is further used to provide sufficient negative concepts for the construction of the word-region alignment loss\, and to complete labels for objects with missing descriptions in captions of image-text pair data. The proposed framework demonstrates strong zero-shot detection performances, e.g., on the LVIS dataset, our DetCLIP-T outperforms GLIP-T by 9.9\% mAP and obtains a 13.5\% improvement on rare categories compared to the fully-supervised model with the same backbone as ours.}

\let\thefootnote\relax\footnotetext{$^*$Equal contribution, $^\dagger$Corresponding author: xu.hang@huawei.com}

\end{abstract}

\section{Introduction}

Most state-of-the-art object detection methods \cite{redmon2016you,ren2015faster,carion2020end,zhu2020deformable}
can only recognize and localize a pre-defined number of categories.
Their detection performance greatly relies on sufficient training
data for each category, which requires expensive and time-consuming
human annotations, especially for the rare classes that can only be distinguished from the expert. 
Even though a great effort has been
made, existing publicly available detection datasets only have a limited number of object categories, for instance, 80 for COCO~\cite{Lin2014}, 365 for Object365~\cite{shao2019objects365}, and 1203 for LVIS~\cite{gupta2019lvis}. However, versatile open-world
object detection is still out of reach mainly for two reasons: a)
Due to the long-tail problem~\cite{gupta2019lvis}, finding sufficient examples for rare
categories is surprisingly challenging; b) Developing a larger detection
dataset requires extremely costly and labor-intensive manual annotation.

On the other hand, image-text pair data are cheap and abundant on
the Internet. Recent vision-language (VL) pre-training
methods (e.g., CLIP~\cite{radford2021learning}, ALIGN~\cite{li2021align})
utilize those data to extend their open-domain capacity and have shown
good zero-shot ability on various downstream classification tasks.
Intuitively, some works \cite{gu2021open,zang2022open,xie2021zsd,du2022learning} try to extend a two-stage detector to an open-world
detector by distilling the learned image embeddings of the cropped
proposal regions from a pre-trained VL model. However, this paradigm requires 
costly feature extraction from cropped images. There also exists discrepancy 
between instance-level features and conventional image-level features extracted 
from VL models. 
\cite{gao2021towards}~proposes a self-training pipeline with pseudo labels~(i.e., noun phrases in caption) generated by a pre-trained VL model~\cite{li2021align} 
while the phrases in caption are always too limited to cover all the objects in an image. 
Moreover, their open-domain ability is determined by the 
performance of the pre-trained VL models.

\begin{figure}[t!]
    \centering
    \includegraphics[width=1.0\linewidth]{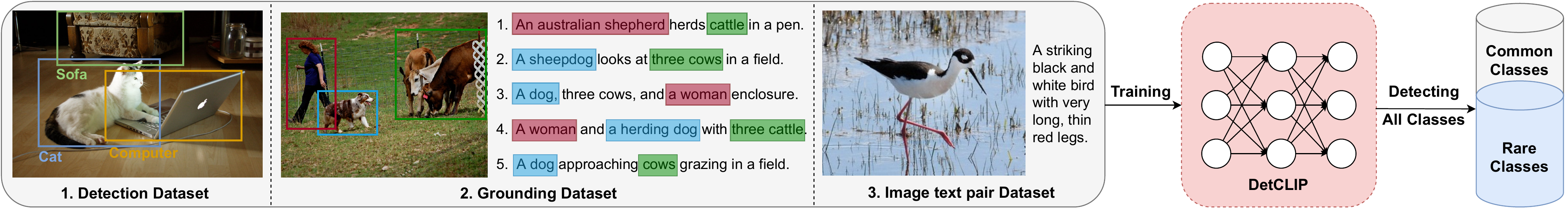}
    \caption{Task definition of open-world detection. In this paper, DetCLIP aims to develop a new pre-training pipeline with datasets from different domains~(including detection dataset, grounding dataset and image-text pair dataset) as input, to deal with the open-world detection problem. }
    \label{fig:example}
    \vspace{-2.5mm}
\end{figure}

Another line of work, GLIP~\cite{li2021grounded} further proposes a 
grounding formulation of open-world detection by directly utilizing both detection and grounding data for pre-training. Specially for detection data, 
GLIP takes an image and a text prompt sentence that concatenates all the category names as input~(i.e., sequential formulation). 
A text encoder will output features for this sentence and then GLIP aligns them with the region features extracted from the image encoder. 
However, restricted by the 
max length of the input token size of the text encoder, it is difficult 
for GLIP to operate on a large number of categories or extend to more 
detailed description of the categories. Furthermore, full attention 
matrix upon all the categories has to be learned in the text encoder, 
which is unnecessary and low-efficient for detection especially when the size of input 
categories increases.

To alleviate the above problems and further improve the open-domain 
ability, we present DetCLIP, a dictionary-enriched visual-concept 
paralleled pre-training method for open-world detection. 
Note that the ``concepts'' denotes the category names in detection data, and the phrases in grounding and image text pair data.
Specifically, we first design \textit{a novel paralleled concept formulation}
to improve learning efficiency. Instead of feeding the whole prompt text sentence into the text encoder like GLIP, DetCLIP extracts
each concept separately and parallelly feeds them into the text encoder~(i.e., \textit{paralleled formulation}). 
This paralleled formulation allows the model to avoid unnecessary interaction between 
uncorrelated categories/phrases, and produce a longer description
for each concept. By converting detection data, 
grounding data, and image-text pair data into paralleled formulation, 
DetCLIP can be pre-trained under different types of supervision to 
support both localization and open-domain capability. 

\begin{wrapfigure}{r}{0.4\textwidth}
\begin{center}
\includegraphics[width=0.39\textwidth]{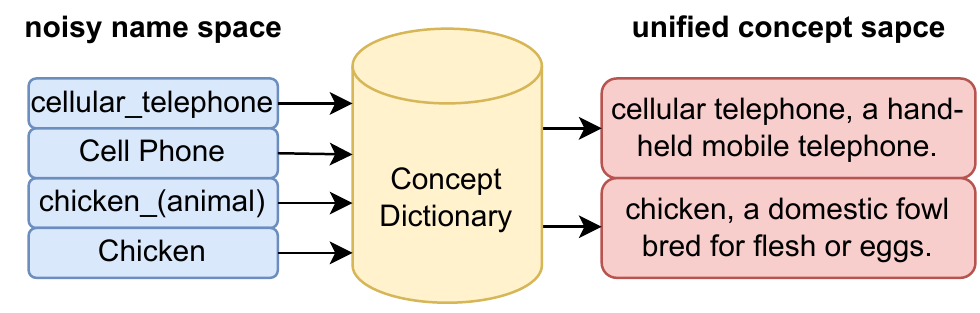}
\end{center}
\caption{DetCLIP aims at unifying the noisy text name space into a joint concept space by a novel concept dictionary.}
\label{fig:intro_dictionary}
\end{wrapfigure}

During pre-training, existing datasets often have a large 
domain gap and difference in their labeling space, e.g., the same 
object concept with different names or hierarchical/inclusive 
structures of different concepts.
However, it is difficult to obtain these implicit relationships among concepts simply by their short names.
In order to form a more unified concept space and provide prior knowledge~(i.e., implicit relationships) for each input concept, we propose \textit{a novel 
concept dictionary} to enrich our prompt text concepts during 
joint pre-training, as shown in Fig.\ref{fig:intro_dictionary}. 
Firstly, we construct the dictionary with concepts extracted from online resources and existing large-scale detection datasets by considering both commonality and coherence.
Based on this concept 
dictionary, DetCLIP can automatically  
enrich the current concept with the concepts and their descriptions existing in the dictionary to facilitate open-domain 
learning.
To alleviate the partial label problem for the grounding and image-text pair data,
DetCLIP further randomly samples concepts from the dictionary as 
negative samples to efficiently pre-train with the alignment loss.
Besides, all concepts in the dictionary are served as the additional category inputs for label completion on the image-text pair data. 
 
The proposed DetCLIP outperforms the state-of-the-art GLIP by large margins on large-scale open-world detection 
benchmark LVIS \cite{gupta2019lvis} with 1203 classes. 
Particularly, DetCLIP-T achieves around 9.9\% mAP 
improvement over GLIP-T on LVIS without utilizing any images in
LVIS during pre-training. 
Compared to the baseline ATSS~\cite{zhang2020bridging} model~(with the same swin-T~\cite{liu2021swin} backbone) trained on LVIS, our DetCLIP-T achieves 2.3\% mAP improvement in the total performance and 13.5\% mAP improvement on the rare classes.

\section{Related Work}

\textbf{Vision-Language Pre-training (VLP)}. Current Vision-Language
Pre-training is a natural extension and development of the successful
pre-train-and-fine-tune scheme in the domains of natural language
processing (NLP) \cite{devlin2018bert,brown2020language} and computer
vision \cite{dosovitskiy2020image} community. Dual-stream methods
such as CLIP \cite{radford2021learning} and ALIGN \cite{li2021align}
have shown great zero-shot classification ability by performing
cross-modal contrastive learning on large-scale image-text pairs from
the Internet. Single-stream approaches \cite{li2019visualbert,kim2021vilt}
directly model the interaction between the visual and textual embeddings
together by a single transformer-based model, which can perform tasks
such as image caption and VQA. Recent approaches such as VLMo
\cite{wang2021vlmo} and BLIP \cite{li2022blip} further explore a
mixed architecture of single-stream and dual-stream models to enable
a unified way of vision-language understanding and generation. However,
those approaches usually focus on whole-image representation learning
and the pre-trained models are usually designed for retrieval/generation
tasks. Current vision-language pre-training approaches can not directly
be applied to object detection task, i.e., a core computer vision
task.

\textbf{General Object Detection.} Object detection is a core problem
in computer vision. CNN-based object detection methods
(one-stage detectors: YOLO \cite{redmon2016you}, SSD \cite{liu2016ssd} and ATSS~\cite{liu2021swin}
and two-stage detectors: Faster R-CNN \cite{ren2015faster} and R-FCN
\cite{dai2016r}) usually use a classifier to map ROI (Region Of Interest)
features into categories. Further improvement of the CNN-based detectors
such as K-Net \cite{zhang2021k} and Panoptic-FCN \cite{li2021fully}
further introduce dynamic kernels and replace the static kernels in
the convolution layers to improve the flexibility
of models. Recent transformer-based methods such as DETR \cite{carion2020end}
and Deformable DETR \cite{zhu2020deformable} try to formulate the
object detection as a set prediction problem that can eliminate post-process NMS. Those methods are constrained
to predefined categories, while our method tries to endow the detector with open-domain
recognition ability which can detect any categories by learning a
wide range of concepts.

\textbf{Zero-shot Object Detection/ Open-vocabulary Object Detection.
}In the early setting of this area, zero-shot object detection aims
to generalize the detector from known categories (training)
to unknown categories (inference). Under this setting, various
works \cite{chen2021knowledge} try to find relationships between existing
categories and unknown categories through pre-trained semantic/text
features \cite{socher2013zero,qiao2016less,reed2016learning,changpinyo2017predicting,elhoseiny2017link}
knowledge graphs \cite{salakhutdinov2011learning,jiang2018hybrid,xu2019reasoning,wang2018zero}
and so on. However, the evaluation under this setting is not general
enough since people usually simply split the class name into known/unknown
categories in the single dataset and the transfer learning
is still under similar domain. On the other hand, inspired
by the success of vision-language\textasciitilde (VL) pre-training
methods (e.g., CLIP \cite{radford2021learning}) and their good zero-shot
ability, several methods attempt to perform zero-shot detection on
a wider range of domains by leveraging a pre-trained VL model. For
example, \cite{gu2021open} tries to distill the learned image embeddings
of the cropped proposal regions from CLIP~\cite{radford2021learning}\}
to a student detector. \cite{gao2021towards} proposes a self-training
pipeline which utilizes Grad-CAM~\cite{selvaraju2017grad} and ALBEF
\cite{li2021align}. However, those methods are very slow because
the feature extraction is repeated and the image-level representation
may be sub-optimal for the instance-wise tasks. Recently, GLIP \cite{li2021grounded}
and X-DETR \cite{cai2022x} try to align region and language features
using a dot-product operation and can be trained end-to-end on both grounding
data and detection data. Our method aims to design an open-domain detector
that learns new concepts efficiently and expands domain coverage 
from low-cost data from the Internet. 

\section{The Proposed Approach}

This paper aims to develop a vision-language pre-training pipeline to enhance new concept learning for open-world detection.
To achieve this goal, our DetCLIP leverages a new paralleled vision-concept per-training pipeline~(Sec.\ref{sec:Paralleled Concept Formulation}) for efficient training and a concept dictionary~(Sec.\ref{sec:Concept Dictionary}) to provide external knowledge to automatically enrich the current input concept and alleviate the partial-labeling  problem of the grounding and image-text pair data. 
Our DetCLIP is pre-trained under hybrid supervision from detection data, grounding data and image-text pair data~(Sec.\ref{sec:model architecture}).

\begin{figure}[t!]
    \centering
    \includegraphics[width=1.0\linewidth]{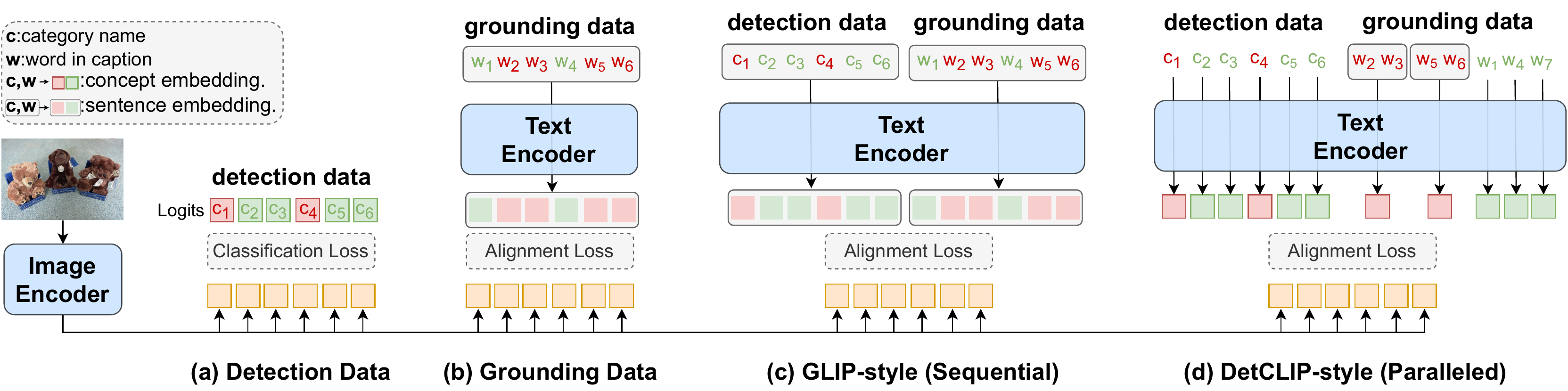}
    \caption{Comparison of different concept formulation strategies. (a) Traditional detection and (b) grounding training utilize different input formulations. (c) To utilize both detection and grounding datasets, GLIP converts the detection data into grounding format by formulating all classes into one sentence as input. (d)  Our DetCLIP introduces a paralleled concept formulation to extract the phrase from the grounding data and regards each phrase as an individual input that is the same as the category name in the detection dataset. Red color denotes that the corresponding concept is in the image while the green indicates the concept is not in the image.}
    \label{fig:architecture_comparison}
    \vspace{-2.5mm}
\end{figure}

\subsection{Paralleled Concept Formulation}
\label{sec:Paralleled Concept Formulation}

A robust open-world detector is required to be trained
with sufficient data covering enough vision concepts. Current available detection datasets lack enough concepts and have limited capacity due to the costly annotation. Leveraging data from other sources, e.g., grounding data and image-text pair data, is a feasible solution to augment the semantic coverage. To enable training with heterogeneous supervisions, we are required to find a unified formulation for different formats of data.

Fig.\ref{fig:architecture_comparison} illustrates a comparison of the different concept formulation strategies that are used for pre-training with different types of data.
As shown in Fig.\ref{fig:architecture_comparison}(a)-(b), traditional detection and grounding training utilize different input formulations. Detection treats each category as a fixed label and aligns region features to the pre-defined label space, while the grounding training leverages the whole caption sentence as input, models the attention between words, and adopts each token embedding in the output embeddings for alignment. To utilize both detection and grounding datasets for pre-training, GLIP~\cite{li2021grounded} converts the detection data to phrase grounding
data, by replacing the object classification logits with the word-region alignment scores, which are associated with a sentence that concatenates
all the category names~(see Fig.\ref{fig:architecture_comparison}(c)), i.e., the text input is [``person, bicycle, car, ... , toothbrush'']. 

We argue that this sequential form is not an effective formulation to model open-world object detection as a vision-language task because it (1) leads to unnecessary interaction between category names in the attention module; and (2) constraints the number of negative samples in contrastive learning due to the limited context length of text input. 
Ablation studies conducted for GLIP show that randomly shuffling the word order in the grounding training data can even bring a slight improvement for the downstream detection task, indicating that the noun phrase is more critical for the detection task, compared to the context information.

\begin{wrapfigure}{r}{5.4cm}
    \centering
    \includegraphics[width=0.9\linewidth, height=3.4cm]{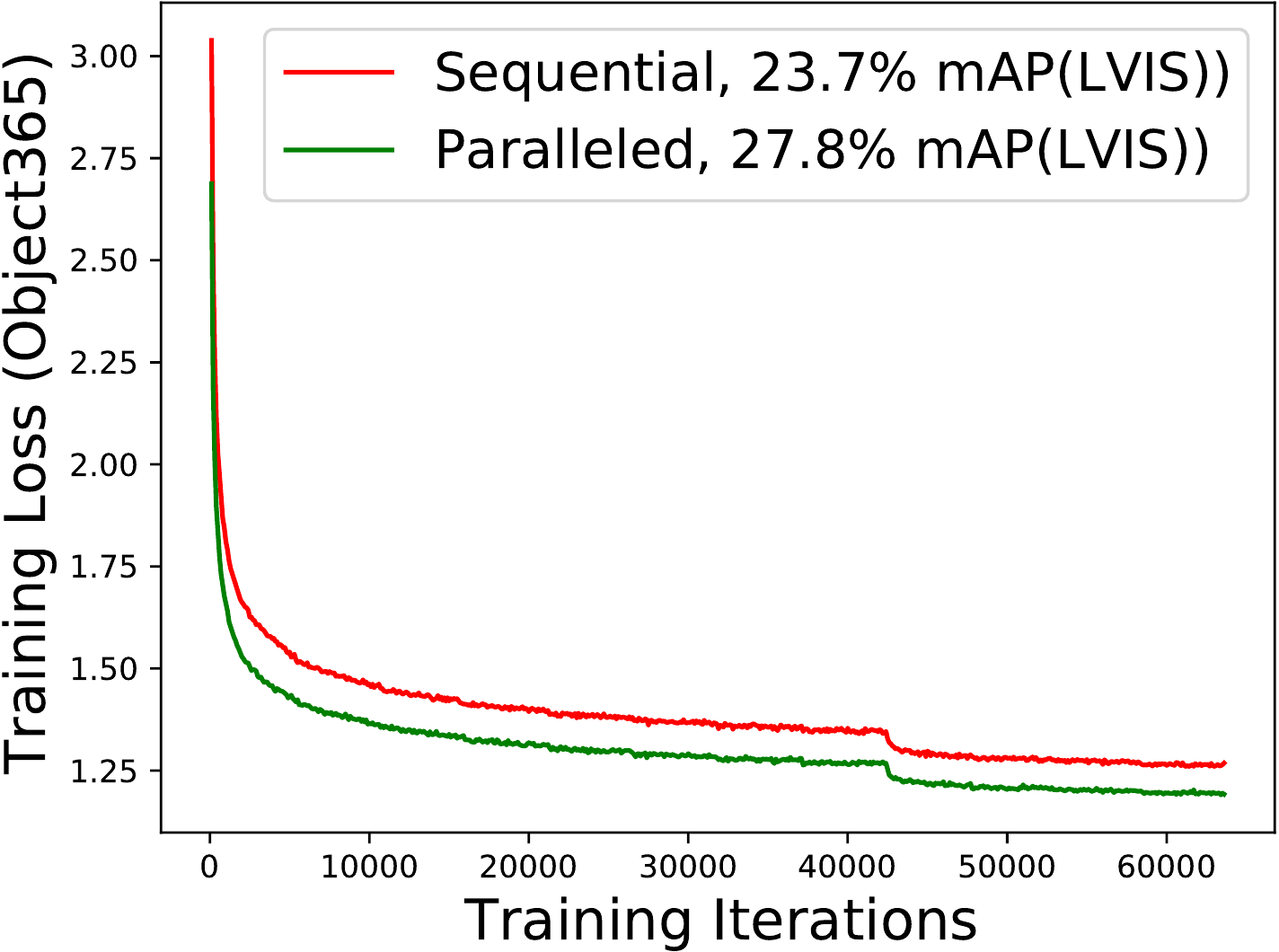}
    \caption{Efficiency comparison between two formulations. Paralleled formulation learns faster and achieves a better performance~(27.8\% v.s. 23.7\%).}
    \label{fig:efficiency_compare}
\end{wrapfigure}

To address the above problem, our DetCLIP introduces a paralleled concept formulation to train with different data sources. 
As shown in Fig.\ref{fig:architecture_comparison}(d), we extract the concept noun phrase for each bounding box and then feed them into the text encoder individually to obtain the corresponding text embedding.
In our paralleled design, context information is removed and model directly learns language features from each separate concepts, which is a more straightforward modeling for the detection task and improves the learning  efficiency (see Fig.\ref{fig:efficiency_compare}).
Furthermore, 
our paralleled design can enable easy expansion
of the augment of category descriptions (see Sec.\ref{sec:Knowledge Enrichment with Concept Dictionary}).
Detailed paralleled formulation for different types of data are as follows:

\textbf{Detection data:} Suppose there are $k$ positive categories in an image, we first pad the category number to $N$ by randomly sampling the negative categories, where $N$ is a predefined number of the concepts for constructing the alignment loss. More specifically,  the text input $P$ can be formulated as $\{p_n\}_{n=1}^N$, where $p_n$ represents for the $n$-th category name, e.g.,\\

\centerline{$P = $ [``person'', ``bicycle'', ``car'', ... , ``toothbrush''].}
Then, we feed $N$ category names as separate sentences into the text encoder and use the embedding of [end of sentence] token as the text embedding for each category. The $N$ text embeddings are then concatenated and matched with ground truth bounding boxes to construct the alignment loss.

\textbf{Grounding data:} We extract positive phrases for bounding boxes (provided by the grounding annotations), and drop other words in the caption. To align the input format with detection data, we also pad the category number to $N$ by sampling negative categories from our proposed concept dictionary (see Sec.\ref{sec:Knowledge Enrichment with Concept Dictionary}). An example of input $P$ for grounding data can be:\\

\centerline{$P = $ [``a woman'', ``a herding dog'', ``three cattle'', ``$neg_1$'', ... , ``$neg_m$''],}
where $neg_1$ to $neg_m$ are sampled negative categories from the constructed concept dictionary. Then similar steps with detection data can be conducted to construct the alignment loss.

\textbf{Image-text pair data:} 
Image-text pair data only contains images and the corresponding captions, while without any annotated bounding boxes. To obtain object-level dense labels, we first use a pre-trained class-agnostic region proposal network~(RPN) to extract object proposals. Then a powerful pre-trained vision-language model like CLIP~\cite{radford2021learning} or FILIP~\cite{yao2021filip} is utilized to assign pseudo labels for the proposals (see Appendix for more details.). 
To tackle the concept-missing problem in the captions of image-text pair data, instead of using noun phrases in the captions as the candidate category names following~\cite{li2021grounded, gao2021towards}, we propose to assign open-domain categories to the object proposals via constructing a large-scale concept dictionary (see Sec.\ref{sec:Knowledge Enrichment with Concept Dictionary}).
After obtaining object-level dense pseudo labels, we convert the image-text pair data to the same format as the detection and grounding data, and then a similar training procedure can be applied .

\subsection{Concept Dictionary\label{sec:Concept Dictionary}}

In this work, we try to build an open-world object detector that can cover a wide range of concepts and can be applicable to different types of datasets. However, the existing
detection/grounding/image-text-pair datasets have a very large domain gap
and difference in their labeling space. For example, a boy can be
annotated as ``man'', ``child'' or ``people'' in the different
datasets. Moreover, there often exists hierarchical/inclusive
relationships between different concepts.
Knowing these implicit relationships can effectively facilitate the pre-training, however it is also clearly challenging to discover these relationships with only a limited set of concept names.

Therefore, we propose to build a large-scale concept dictionary to form a unified
concept space for different data sources and explicitly provide the useful relationships between various concepts through definitions.
 For example, the car is defined as ``a motor vehicle with four wheels usually propelled by an internal combustion engine'', and the motorcycle is defined as ``a motor vehicle with two wheels and a strong frame''.  By aggregating these definitions, we can conclude that the car and the motorcycle are both motor vehicles with a difference in the number of wheels.

\subsubsection{Constructing the Concept Dictionary}

To construct a unified concept dictionary $O$, we collect concepts from multiple sources: (1) noun phrases extracted from the large-scale image-text pair dataset, i.e., YFCC100m; (2) category names of existing public detection datasets (e.g., Objects365~\cite{shao2019objects365}, OpenImages~\cite{kuznetsova2020open}); (3) object names from the manually-collected concept database, i.e., Things~\cite{hebart2019things}). 
For object names from the detection datasets and Things, we directly add them into the dictionary after deduplication.
For noun phrases extracted from the YFCC, we filter out the concepts that appear less than a frequency of $100$ or without definition in WordNet~\cite{Miller1998wordnet} to ensure the commonality.
Our concept dictionary $O$ is then constructed
by concatenating each word $o_l$ with its definition $def_l$ in WordNet: $O=\{o_l:def_l\}_{l=1}^L, $
where $L$ is the number of concepts in our dictionary. 
The constructed dictionary covers about 14k concepts along with their definitions, and can be updated by directly adding new concepts and corresponding definitions. Based on the constructed concept dictionary, we then propose 2 techniques utilizing it to boost the pre-training in the following section. 

\subsubsection{Knowledge Enrichment with Concept Dictionary}
\label{sec:Knowledge Enrichment with Concept Dictionary}

\begin{figure}[t!]
    \centering
    \includegraphics[width=1.0\linewidth]{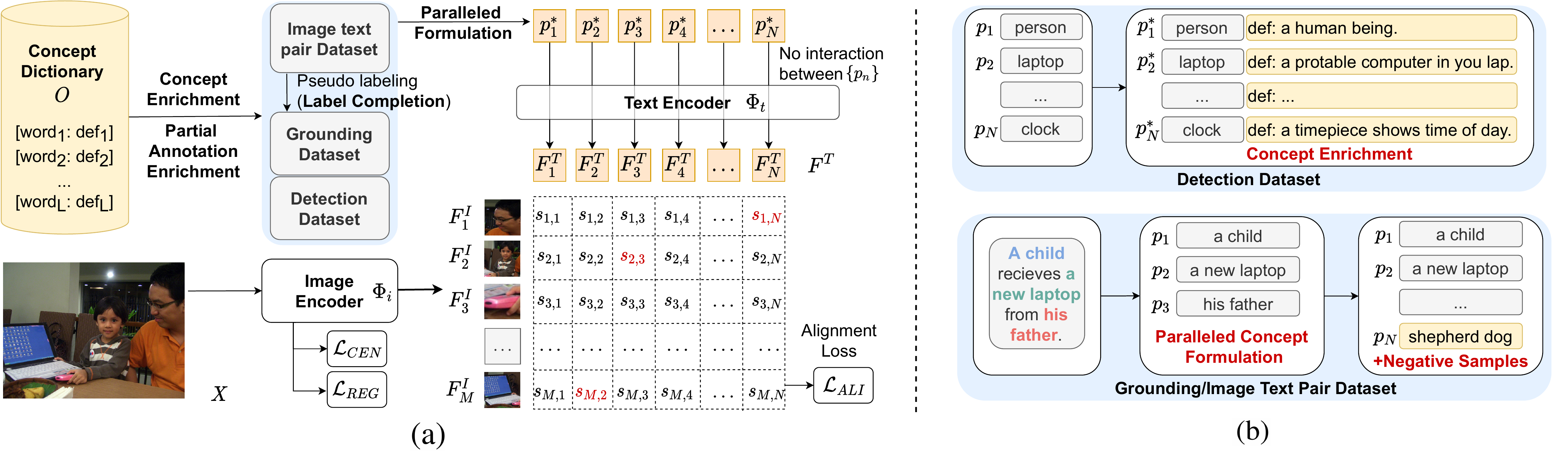}
    \caption{Overall architecture and the details of utilizing concept dictionary $O$. (a) DetCLIP contains an image encoder $\Phi_i$ to obtain region features $F^I$ and a text encoder $\Phi_t$ to get embeddings $F^T$ for each enriched concept $p^*_n$. Then the region-concept alignment loss $\mathcal{L}_{ALI}$ is performed. Note that the box regression loss $\mathcal{L}_{REG}$ is only adopted on detection datasets. (b) A concept dictionary $O$ is introduced to enrich the current concept with prior knowledge and provide negative category samples for construction of the alignment loss. }
    \label{fig:overall_architecture}
\end{figure}

\textbf{Concept Enrichment.}  
Based on the designed concept dictionary, we first retrieve the definition for each input concept to provide the prior knowledge~(see Fig.\ref{fig:overall_architecture}(b)).
During pre-training, for each concept $p_n$ in
the training set, we can directly use its definition if $p_n$ is included in the dictionary $O$. 
If we cannot find a direct match in $O$, we will try to locate the most related concept in $O$ by calculating a similarity matrix $S'\in \mathbb{R}^{L}$.
The $S'$ is calculated via the dot-product of the embeddings from a pre-trained text encoder such as FILIP~\cite{yao2021filip} with $p_n$ and all concept names $\{o_l\}_{l=1}^L$ as input. Then we can find the most related concept $\{o_{l^*}$, where ${l^*}=argmax_l(S'(l))\}$ in the dictionary, to retrieve an
approximate definition $def_{{l^*}}$. 
The $p_n$ is then enriched with the retrieved definition and reformatted as $\{p^*_n\} = \{p_n, def_{l^*}\}$. An example of the enriched text input $P^*=\{p^*_n\}_{n=1}^N$ is:
\begin{center}
{$P^* = $ [``person, a human being.'', ``bicycles, a wheeled vehicle that has two wheels and is moved by foot pedals.'', ... , ``toothbrush, small brush has long handle used to clean teeth.'']}
\end{center}
\textbf{Partial Annotation Enrichment.}
In the grounding or image-text pair data, only main objects that people care about are labeled in the caption, which is known as the partial labeling problem.
Compared with standard detection datasets which have sufficient positive and negative classes for each image, pre-training with grounding and image-text pair datasets encounters two severe issues: 1) lack of annotations of negative concepts for learning discriminative concept embeddings; 2) lack of annotations of partial positive concepts to efficiently train the model.
For the first problem, DetCLIP randomly samples the concepts in the constructed dictionary $O$ as the negative concepts to construct the alignment loss, instead of directly padding empty inputs~(Fig.\ref{fig:overall_architecture}(b)).
Note that since the number of concepts in the dictionary $O$ is large~(i.e., about 14k), the probability that the sampled concepts are indeed in the image is extremely small.
For the second problem, to perform label completion on image-text pair data during pseudo labeling, we add all the concepts in dictionary $\{o_l\}_{l=1}^L$ as the additional category inputs, instead of using the original noun phrase in the caption to calculate the similarity matrix. 
Therefore, the concepts shown in the image while not in the caption can also be labeled and then get pre-trained.
An illustration is also shown in Fig.\ref{yfcc_plabel} to qualitatively verify the effectiveness of label completion. 

\subsection{Model Architecture/Training Objective}
\label{sec:model architecture}

As shown in Fig.\ref{fig:overall_architecture}, the basic architecture of DetCLIP contains an image encoder $\Phi_i$ to generate the region features $F^I\in \mathbb{R}^{M \cdot D}$ from the input image $X$, and a text encoder $\Phi_t$ to obtain the text embeddings $F^T\in \mathbb{R}^{N \cdot D}$ for the concepts in $P^*$, where 
$M$, $N$ denote for the number of extracted regions and input concepts, respectively.
Then the alignment loss is constructed by calculating the alignment score $S\in \mathbb{R}^{N \cdot M}$ for all region-text pairs. 
\begin{equation}
\label{eq:extract feature}
F^I = \Phi_i(X), F^T = \Phi_t(P^*), S = \langle{F^I},{\rm{Transpose}(F^T)}\rangle
\end{equation}

With the ground-truth alignment matrix $G\in \mathbb{R}^{N \cdot M}$, the whole training objective $\mathcal{L}$ can be written as:

\begin{equation}
\label{eq:whole loss}
\mathcal{L} = \mathcal{L}_{ALI}(S, G) + \alpha \cdot \mathcal{L}_{CEN} + \beta \cdot \mathcal{L}_{REG},
\end{equation}

where $\mathcal{L}_{ALI}$, $\mathcal{L}_{CEN}$, and $\mathcal{L}_{REG}$ denote the alignment loss, the centerness loss, and the regression loss, respectively. $\alpha$ and $\beta$ represent the weight factor for $\mathcal{L}_{CEN}$ and $\mathcal{L}_{REG}$, respectively.
Following GLIP~\cite{li2021grounded}, we adopt the ATSS~\cite{zhang2020bridging} detector as our image encoder. We use the sigmoid focal loss~\cite{lin2017focal} for $L_{ALI}$, the sigmoid loss for $L_{CEN}$, and the GIoU loss~\cite{rezatofighi2019generalized} for $L_{REG}$.

\section{Experimental Results}
\label{subsec:Implementation Details}

\begin{table}[t!]
\begin{centering} 
\caption{\label{tab:Preliminary-Experiment-1:}{Zero-shot performance on LVIS~\cite{gupta2019lvis} minival datasets. AP$_r$ / AP$_c$ / AP$_f$ indicate the AP
values for rare, common, frequent categories, respectively. `DH' and `F' in GLIP~\cite{li2021grounded} baselines stand for the dynamic head \cite{dai2021dynamic} and cross-modal fusion, respectively. Baselines with * are implemented with our code-base.
GoldG+ denotes the GoldG plus the COCO~\cite{Lin2014} caption dataset.
}}
\label{tab:LVIS+13dataset}
\par\end{centering}
\begin{center}
\begin{small}
\begin{sc}
\begin{tabular}{c|cc|cccc}
\toprule 
\multirow{2}{*}{Model} & \multirow{2}{*}{Backbone} & \multirow{2}{*}{Pre-Train Data} & \multicolumn{2}{c}{LVIS} \tabularnewline
 &  &  & {AP} & {AP$_r$ / AP$_c$ / AP$_f$} \tabularnewline
  \midrule
  {Mask-RCNN\cite{he2017mask}*} & {Swin-T}  & {LVIS} & {34.1} & {19.1 / 34.0 / 37.0}  \tabularnewline
 {ATSS\cite{zhang2020bridging}*} & {Swin-T} & {LVIS} & {33.6} & {19.7 / 32.4 / 37.2} \tabularnewline
 {ATSS\cite{zhang2020bridging}*} & {Swin-L} & {LVIS} & {43.9} & {30.6 / 43.7 / 46.3} \tabularnewline
 \midrule
 {MDETR\cite{kamath2021mdetr}} & {RN101} & {GoldG+} & {24.2} & {20.9 / 24.3 / 24.2} \tabularnewline
 {GLIP-T(A)\cite{li2021grounded}} & {Swin-T+DH+F} & {O365} & {18.5} & {14.2 / 13.9 / 23.4} \tabularnewline
 
  {GLIP-T(C)\cite{li2021grounded}} & {Swin-T+DH+F} & {O365,GoldG} & {24.9} & {17.7 / 19.5 / 31.0} \tabularnewline
   
 {GLIP-T\cite{li2021grounded}} & {Swin-T+DH+F} & {O365,GoldG,Cap4M} & {26.0} & {20.8 / 21.4 / 31.0} \tabularnewline
 {GLIP-L\cite{li2021grounded}} & {Swin-L+DH+F} & {4ODs,GoldG,Cap24M} & {37.3} & {28.2 / 34.3 / \textbf{41.5}} \tabularnewline
 
  {GLIPv2-T\cite{zhang2022glipv2}} & {Swin-T+DH+F} & {O365,GoldG,Cap4M} & {29.0} & { \ \ \ - \ \ / \ \ \  - \ \ \ / \ \ \ -\ \ \ } \tabularnewline
 
  \midrule
{DetCLIP-T(A)} & {Swin-T} & {O365} & {28.8} & {26.0 / 28.0 / 30.0}\tabularnewline
{DetCLIP-T(B)} & {Swin-T} & {O365, GoldG} & {34.4} & {26.9 / 33.9 / 36.3}\tabularnewline
\rowcolor{mypink}{DetCLIP-T} & {Swin-T} & {O365, GoldG, YFCC1m} & {\textbf{35.9}} & {\textbf{33.2} / \textbf{35.7} / \textbf{36.4}} \tabularnewline
\midrule
\rowcolor{mypink} {DetCLIP-L} & {Swin-L} & {O365, GoldG, YFCC1m} & {\textbf{38.6}} & {\textbf{36.0 }/\textbf{ 38.3 }/\textbf{ 39.3}} \tabularnewline
\bottomrule

\end{tabular}
\end{sc}
\end{small}
\end{center}
\vspace{-4mm}
\end{table}

\textbf{Implementation Details.} We pre-train all the models based on Swin-Transformer \cite{liu2021swin} backbones with 32 GPUs.
AdamW optimizer \cite{kingma2014adam} is adopted and batch size is set to 128.
The learning rate is set to 2.8x10$^{-4}$ for the parameters of the visual backbone and detection head, and 2.8x10$^{-5}$ for the language backbone.
Without otherwise specified, all models are trained with 12 epochs and the learning rate is decayed with a factor of 0.1 at the 8-{th} and the 11-{th} epoch.
The max token length for each input sentence is set to 48.
The number of the concepts $N$ in text input $P$ is set to 150 and the number of region features $M$ is determined by the feature map size and the number of pre-defined anchors. 
The loss weight factors $\alpha$ and $\beta$ are both set to 1.0.
The training of DetCLIP-T with Swin-T backbone tasks 63 hours when 32 GPUs are used. MMDetection~\cite{mmdetection} code-base is used.

\textbf{Training Data.}
Our model is trained with a hybrid supervision from different kinds of data, i.e., detection data, grounding data, and image-text pair data. More specifically, for detection data, we use a sampled Objects365 V2 ~\cite{shao2019objects365} dataset (denoted as O365 in the following sections) with 0.66M training images. Here we do not use the whole dataset since training with sampled data is more efficient and suffices to demonstrate the effectiveness of our method. For grounding data, we use gold grounding data (denoted as GoldG) introduced by MDETR \cite{kamath2021mdetr}. Moreover, following GLIP \cite{li2021grounded}, we remove the training samples contained in LVIS~\cite{gupta2019lvis} dataset for fair zero-transfer evaluation, which results in 0.77M training data. For image-text pair data, we perform object-level dense pseudo labeling on YFCC100m \cite{thomee2016yfcc100m} dataset with a pre-trained CLIP \cite{radford2021learning} model, and sample a subset of 1M training images from the results containing objects with similarities scores above a given threshold. Finally, our training set contains a total of 2.43M images. Compared to GLIP's 27M training data, \textit{DetCLIP only uses less than 10\% data, but achieves better results}. More details are in Appendix.

\label{subsec:Benchmark Settings}
\textbf{Benchmark Settings. }We evaluate our method mainly on LVIS \cite{gupta2019lvis} which contains 1203 categories. 
Following GLIP \cite{li2021grounded} and MDETR \cite{kamath2021mdetr}, we evaluate on the 5k minival subset and report the zero-shot \textit{fixed} AP~\cite{dave2021evaluating} for a fair comparison. We do not focus the performance on COCO \cite{Lin2014} since it only contains 80 common categories that are fully covered by the training dataset Objects365~\cite{shao2019objects365}, which may not sufficient to reflect generalization ability of a model in the open-domain detection setting.
To further study the generalization ability of our method, following GLIP~\cite{li2021grounded}, we also evaluate the averaged AP on other 13 downstream detection datasets published on Roboflow\footnote{\url{https://public.roboflow.com/object-detection}}.

\subsection{Open-world Detection Results\label{subsec:Open-world Detection Results}}
\label{exps:detection results}

We train our DetCLIP with two backbones, i.e., swin-T~\cite{liu2021swin} and swin-L. Distinct from GLIP~\cite{li2021grounded}, which introduces additional heavy modules like Dynamic Head \cite{dai2021dynamic} and cross-modal fusion, we directly adopt the vanilla ATSS \cite{zhang2020bridging} as our vision encoder to keep our architecture as neat as possible. Following GLIP, for Swin-T backbone, we also train 3 versions of models, i.e., DetCLIP-T(A), DetCLIP-T(B) and a complete version  DetCLIP-T, which differ in using different training data. 

Table \ref{tab:LVIS+13dataset} reports the results on LVIS~\cite{gupta2019lvis}. Results with LVIS as pre-training data stand for the fully-supervised models trained with annotated data. With our proposed paralleled formulation, introducing more training data from different sources can consistently improve the performance. I.e., comparing  DetCLIP-T(A) trained with only detection data with DetCLIP-T trained with additional grounding and image-text pair data, we can observe a considerable performance gain (28.8\% AP v.s. 35.9\% AP). Besides, befitting from the proposed effective framework, our DetCLIP models outperform their GLIP counterparts by a large margin, i.e., DetCLIP-T(A) (resp. DetCLIP-T) surpasses GLIP-T(A) (resp. GLIP-T) by 10.3\% (resp. 9.9\%). {Besides, DetCLIP-T also significantly outperforms GLIPv2-T by 6.9\%.} Note that our models are more lightweight (without heavy DyHead and cross-modal fusion) and trained with fewer epochs (our 12 vs. GLIP's 24) and much less data. In addition, our DetCLIP-T's zero-shot performance \textit{even beats the fully-supervised model with the same backbone} by utilizing weak-annotated data like image-text pair data.
Results on LVIS full validation dataset can refer to the Appendix. 

\textbf{Efficiency Comparison.}
To demonstrate the efficiency of our proposed DetCLIP, we directly compare the training and inference speed of DetCLIP-T with the GLIP-T~\cite{li2021grounded} in Table~\ref{tab:Efficiency comparison}. 
\begin{wraptable}{r}{7.0cm}
\caption{Efficiency comparison on LVIS.}
\label{tab:Efficiency comparison}
\begin{center}
\begin{small}
\begin{sc}
\begin{tabular}{c|cc}
\toprule
Model& Training & Inference \\
\midrule
GLIP-T&~10.7K gpu hrs & 0.12 FPS \\
\rowcolor{mypink} DetCLIP-T& \textbf{2.0K} gpu hrs & \textbf{2.3} FPS \\
\bottomrule
\end{tabular}
\end{sc}
\end{small}
\end{center}
\end{wraptable}
With the same setting of training with 32 V100 GPUs, the total training time for GLIP-T is about 10.7K GPU hours~(5X than us) due to its heavy backbone and more image-text pair training data.
On the other hand, DetCLIP-T achieves the 2.3 FPS~(0.43 s/image) on a single V100 when performing inference on LVIS~\cite{gupta2019lvis}, while GLIP-T can only achieve 0.12 FPS~(8.6 s/image).
With much better training and inference efficiency, DetCLIP-T can still outperform GLIP-T 9.9\% on LVIS.

\begin{figure}[t!]

    \centering
    \includegraphics[width=\linewidth]{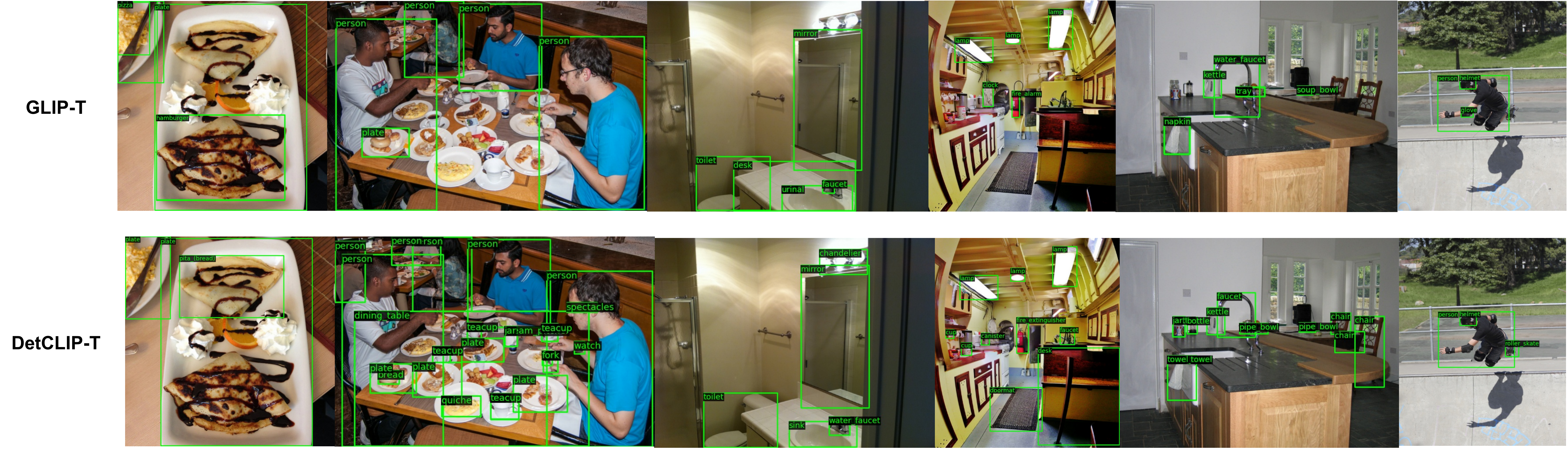}
    \caption{Qualitative prediction results from GLIP-T~\cite{li2021grounded} (first row) and DetCLIP-T (second row) on LVIS~\cite{gupta2019lvis} dataset. DetCLIP-T can produce more complete and precise predictions than GLIP-T.}
    \label{fig:Qualitative Visualizations.}
    \vspace{-3mm}
\end{figure}

\textbf{Qualitative Visualizations.}
We illustrate the bounding box predictions on LVIS~\cite{gupta2019lvis} dataset from DetCLIP-T and GLIP-T~\cite{li2021grounded} model in Fig.\ref{fig:Qualitative Visualizations.}. We can observe that by adopting the paralleled concept formulation and the external knowledge from the concept dictionary, our DetCLIP can outperform GLIP both on the accuracy and completeness of the predicted labels, especially on the rare classes. 

\subsection{Ablation Studies}\label{subsec:Ablation Studies}

\begin{table}[t!]
\caption{Ablation Studies of different components on LVIS~\cite{gupta2019lvis} minival and other 13 detection datasets~("13 DATA"). Where P.F, C.E, N.S, L.C stand for the paralleled formulation, concept enrichment, negative samples and label completion. The latter 3 techniques are realized by the proposed concept dictionary. Numbers in parentheses under LVIS column are  AP$_r$/AP$_c$/AP$_f$, respectively.}
\label{tab:ablation}
\begin{center}
\begin{small}
\begin{sc}
\begin{tabular}{c|cccc|cc}
\toprule
 Pre-training Data& P.F & C.E. & N.S  & L.C & LVIS  & 13 data \\
\midrule
\multirow{3}{*}{O365\cite{shao2019objects365}}&\xmark &  \xmark  &  \xmark  & \xmark& 23.7 (16.6/20.5/27.7)&  30.1 \\
&\cmark& \xmarkg   &  \xmarkg  &\xmarkg &27.8 (22.2/26.8/29.7) &30.7  \\
&\cmark&\cmark  & \xmarkg &\xmarkg & 28.8 (26.0/28.0/30.0)& 33.8\\
\midrule
\multirow{4}{*}{O365\cite{shao2019objects365}, GOLDG\cite{Krishna2016}}&\cmark&\xmark&\xmarkg  &\xmarkg & 28.2 (21.6/25.0/32.2) & 33.7\\
&\cmark&\cmark&\xmarkg  &\xmarkg & 32.2 (26.4/30.3/34.9)& 35.9\\
&\cmark&\xmark & \cmark&\xmarkg & 30.3 (22.6/27.4/34.2)& 36.3 \\
&\cmark&\cmark & \cmark&\xmarkg & 34.4 (26.9/33.9/36.3)& 38.8 \\
\midrule
\multirow{2}{*}{O365\cite{shao2019objects365}, GOLDG\cite{Krishna2016}, YFCC\cite{thomee2016yfcc100m}} &
\cmark&\cmark & \cmark &\xmarkg & 35.5(32.8/34.9/\textbf{36.6})  &    39.9  \\
&
\cellcolor{mypink}\cmark&\cellcolor{mypink}\cmark & \cellcolor{mypink}\cmark &\cellcolor{mypink}\cmark & \cellcolor{mypink}\textbf{35.9}(\textbf{33.2}/\textbf{35.7}/36.4)&      \cellcolor{mypink}\textbf{43.3}  \\
\bottomrule
\end{tabular}
\end{sc}
\end{small}
\end{center}
\end{table}

\begin{table}[h!]
\begin{centering} 
\caption{{ The impact of the scale of the concept dictionary.    The numbers in parentheses indicate the size of the corresponding concept dictionary. The first row means without using the concept dictionary. Zero-shot performances on LVIS~\cite{gupta2019lvis} are reported.
}}
\label{tab:Different_Size_of_the_Dictionary}
\par\end{centering}
\begin{center}
\begin{small}
\begin{sc}
\resizebox{\textwidth}{!}{
\begin{tabular}{c|c|cccc}
\toprule 
\multirow{2}{*}{Model} & \multirow{2}{*}{Concept Dictionary} & {LVIS minival} & {LVIS val} \tabularnewline
 &  & {AP} ({AP$_r$ / AP$_c$ / AP$_f$}) & {AP} ({AP$_r$ / AP$_c$ / AP$_f$}) \tabularnewline
  \midrule
\multirow{4}{*}{DetCLIP-T(B)} & {/} & 28.2 (21.6 / 25.0 / 32.2) & 20.9 (15.3 / 17.5 / 27.1) \tabularnewline
 & {O365 (\textasciitilde 0.36K)} & 27.8 (22.3 / 23.7 / 32.4) & 21.6 (19.3 / 18.7 / 25.8)  \tabularnewline
 & {O365+Things (\textasciitilde 1.9K)} & 28.1 (20.8 / 24.8 / 32.4) &  20.5 (14.0 / 17.0 / 27.2) \tabularnewline
& {Detection + Image-text (\textasciitilde 14K)} &  \textbf{34.4} (\textbf{26.9} / \textbf{33.9} / \textbf{36.3}) &  \textbf{27.2} (\textbf{21.9} / \textbf{25.5} / \textbf{31.5})  \tabularnewline
\bottomrule
\end{tabular}
}
\end{sc}
\end{small}
\end{center}
\end{table}

Table \ref{tab:ablation} studies the effectiveness of two core components of DetCLIP, i.e., the paralleled formulation and the knowledge enrichment with concept dictionary. The first row stands for our implementation of a GLIP-A~\cite{li2021grounded} model, which is modeled with sequential formulation and uses only detection data for training. Due to the implementation discrepancy, our version can achieve 23.7\% zero-shot AP on LVIS~\cite{gupta2019lvis}, which is higher than the official's 18.5\% and serves as a stronger baseline. 

First, applying the paralleled formulation can bring significant improvements (row 2), boosting the performance to 27.8\%. This indicates that paralleled formulation is much more effective than sequential formulation for modeling object detection as a vision-language task. However, directly applying the same approach to a larger scale dataset with heterogeneous supervisions, e.g., detection plus grounding, can hurt the performance (row 4). We speculate this is because paralleled formulation weakens the interaction between text concepts, resulting in the model's inability to effectively construct the connections between semantic-related concepts. Therefore, we introduce word definitions to the class names to help bridge relationships between different concepts, which boosts the performance to 32.2\% (row 5). Sampling negative categories from the concept dictionary also helps better utilize grounding data, improving the performance to 34.4\% (row 7). Further introducing image-text pair datasets like YFCC~\cite{thomee2016yfcc100m} can bring substantial improvement for rare categories (row 8), while utilizing concept dictionary for label completion during pseudo labeling finally improves the overall AP to 35.9\% (row 9). A similar performance pattern is also observed on 13 downstream detection datasets. 

\begin{figure}
    \centering
    \includegraphics[width=\linewidth]{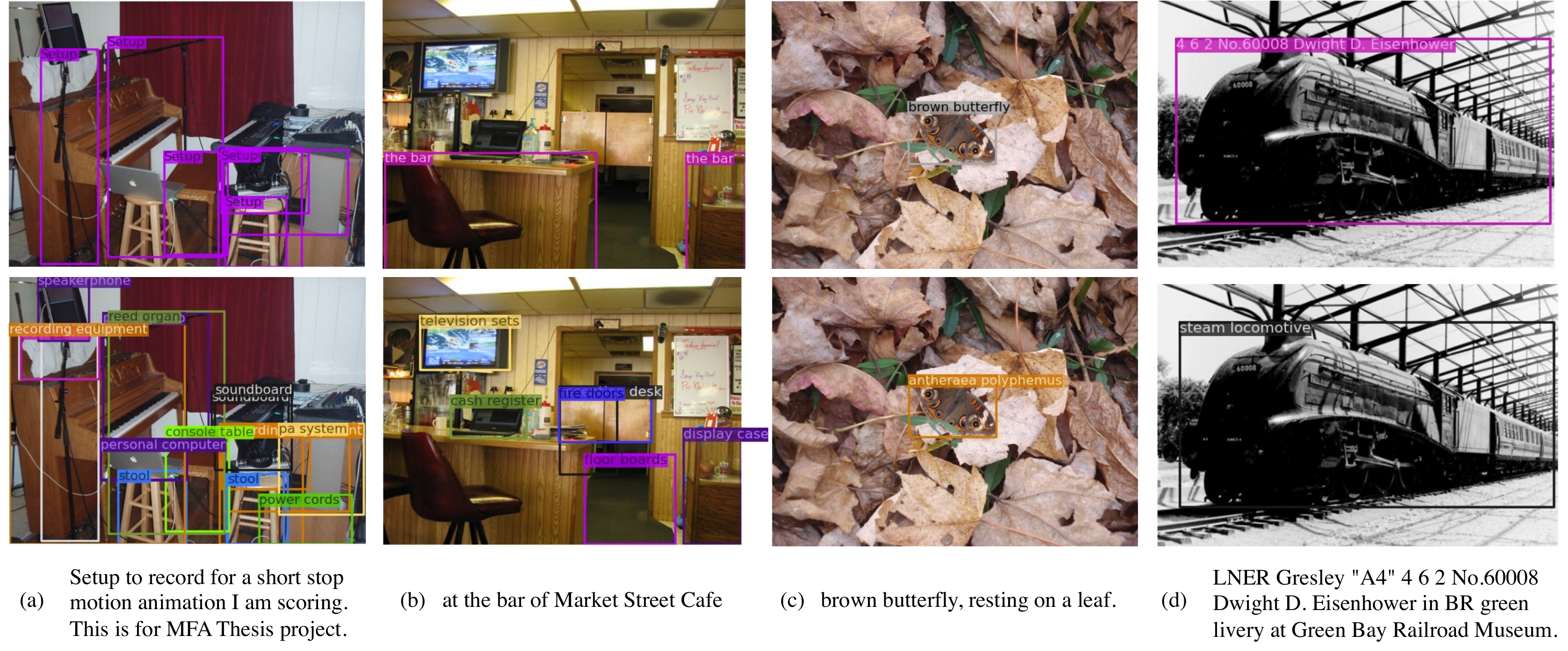}
    \caption{Visualization of pseudo label results on YFCC~\cite{thomee2016yfcc100m} dataset. The texts below the images are the corresponding captions. Top row: labeling results using the original captions; bottom row: labeling results with the designed concept dictionary. It helps produce higher-quality pseudo labels.}
    \label{yfcc_plabel}

\end{figure}
\begin{figure}
    \centering
    \includegraphics[width=0.9\linewidth]{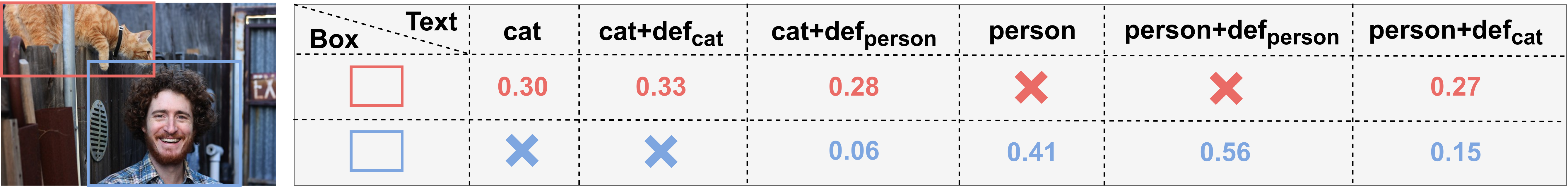}
    \caption{Alignment scores when using text inputs with different definition settings. Cross mark means the scores are below the threshold. $def_x$ means we use the definition of category x. Results show that the definition can play an important role in classifying objects. }
    \label{fig:def_fusion.}
    \vspace{-4mm}
\end{figure}

\textbf{Impact of Concept Dictionary's Size.} 
To study the impact of the scale of the proposed concept dictionary, we build three concept dictionaries with different sizes by using: (1)  class names from Objects365 \cite{shao2019objects365}; (2) class names from Objects365 + Things \cite{hebart2019things}; (3)  class names in (2) plus noun phrases extracted from YFCC100m~\cite{thomee2016yfcc100m}. We equip our DetCLIP-T(B) with these three concept dictionaries and compare their performance in Table \ref{tab:Different_Size_of_the_Dictionary}. It can be seen that using a small size dictionary, e.g., Objects365 + Things, can even bring a performance drop compared to without the dictionary, while scaling up the dictionary with nouns from YFCC can significantly improve the performance. We speculate this is because a large dictionary can provide rich negative concepts for grounding data, encouraging the model to learn more discriminative features.

\textbf{Pseudo labels with concept dictionary.} Fig.\ref{yfcc_plabel} visualizes the results of label completion via concept dictionary. Multiple effectiveness can be observed. E.g., cases (a) and (b) demonstrate concept dictionary contributes to labeling bbox with the category names not shown in the captions. In (c), a finer-grained pseudo label can be produced, i.e., `antheraea polyphemus' v.s. the original `brown butterfly', which helps the learning of rare categories. In (d), label noise in the caption is alleviated.

\textbf{Importance of concept enrichment.} To directly illustrate how the category definition helps the model achieve better detection performance, we conduct the experiments to infer with different text inputs with DetCLIP-T utilized.
We compare the three different cases, i.e., no definition, true definition and false definition (use definition of another category), and report the alignment score for different objects in Fig.\ref{fig:def_fusion.}.
Observations are: (1) True definition helps the model to better determine the category~(i.e., increases the confidence score); (2) Wrong definition may confuse the model and bring the false positive samples; (3) Both the category name and definition matter to the model.

\section{Conclusion}
In this paper, we propose a novel open-world detection pre-training framework named DetCLIP, aiming at improving the open-domain ability and learning efficiency of the open-world detector.
By unifying three kinds of supervision via paralleled concept formulation, our DetCLIP can learn from different domains and enhance the training efficiency.
We further propose a concept dictionary module to improve the discovery and coverage of the novel knowledge by importing external knowledge. 
The proposed usages of the concept dictionary achieves better open-world detection result in terms of both common and rare categories on LVIS.
Experiments on multiple downstream detection datasets suggest that our DetCLIP is more powerful than current SOTA open-world detectors such as GLIP~\cite{li2021grounded}.

\noindent \textbf{Acknowledgements} We gratefully acknowledge the support of MindSpore\footnote{\url{https://www.mindspore.cn/}}, CANN~(Compute Architecture for Neural Networks) and Ascend AI Processor used for this research.

\section*{Appendix for DetCLIP: Dictionary-Enriched Visual-Concept Paralleled Pre-training for Open-world Detection}

\appendix


\section{Negative Impacts and Limitations}

\textbf{Potential Negative Social Impact.} Our method has no ethical risk on dataset usage and privacy violation since all the benchmarks are publicly available and transparent.

\textbf{Limitations and Future Works.} The localization ability of the region proposals is still limited by the annotation of the bounding box. More weakly-supervision can be included to learn from image-text pairs. 
Furthermore, although we prove the effectiveness of our method on the web-collected dataset YFCC~\cite{thomee2016yfcc100m},  we expect to extend our method to larger image-text pair datasets from the Internet.

\section{Dataset Details}
In this section, we provide more details of training datasets used in our experiments, which include (1) the approach of generating pseudo detection labels for image-text pair dataset; and (2) a dataset comparison with GLIP~\cite{li2021grounded}.

\textbf{Pseudo Labeling on Image-Text Pair Data.} We use image-text pair data from the web-collected dataset YFCC100m~\cite{thomee2016yfcc100m}. To generate pseudo detection labels for image-text pair data, we first use a Region Proposal Network (RPN) pre-trained on Objects365 to extract object proposals. To ensure the quality of proposals, we filter result bounding boxes with objectness scores below a threshold of 0.3 or region area smaller than 6000. {This operation also helps significantly reduce the number of proposal candidates and accelerates the pseudo-labeling process}.Then a powerful pre-trained CLIP~\cite{radford2021learning} model (ViT-L) is used to predict pseudo class labels for each retained bounding box. To alleviate the partial-label problem, we use concept names from our proposed concept dictionary (Sec.3.2) instead of the raw caption as the text input. Following CLIP, the prompt "a photo of a {category}." is used to pad a category name into the sentence. {Since the proposed dictionary consists of a large number of concepts (i.e., 14k), to accelerate the inference, we pre-compute the text embeddings of all concepts and store them for the later computation.} For each proposal bounding box, we first crop it from the raw image, then resize it to $224\times224$ and feed it into the visual encoder to obtain the visual embedding. We use the cosine similarity as the classification score, which is computed as
\begin{equation}
\label{eq:extract feature_2}
s_i^j = f_\theta^I(R_i)g_\phi^T(c_j)^\top
\end{equation}
where $f_\theta^I$ and $g_\phi^T$ stand for the image and text encoder of the CLIP model; $R_i$ and $c_j$ are $i$-th cropped proposal and $j$-th category, respectively. {Both embeddings are L2-normalized before the similarity calculation.} After category prediction, a second-stage filtering is adopted to drop proposals with a classification score below $0.24$. Finally, we sample 1M images from the results to form our final training  image-text pair data.

\textbf{Training Data Comparison (with GLIP~\cite{li2021grounded}).} Table \ref{tab:dataset_info} compares the training data used by DetCLIP and GLIP. The explanation of each dataset can be found in the table caption. Our DetCLIP-T uses less than half training data compared to GLIP-T, while DetCLIP-L uses \textbf{less than 10\%} training data compared to GLIP-L.

\begin{table}[ht!]
\caption{A training data comparison between DetCLIP and GLIP~\cite{li2021grounded}. Numbers in parentheses indicate the volume of the corresponding dataset. O365-V1 and -V2 are 1st and 2nd version of Objects365 \cite{shao2019objects365} dataset, respectively. 4ODs is a combination of 4 detection datasets, i.e., Objects365~\cite{shao2019objects365}, OpenImages~\cite{krasin2017openimages}, Visual Genome~\cite{Krishna2016} and ImageNetBoxes~\cite{krizhevsky2012imagenet}. GoldG is the grounding data introduced by MDETR\cite{kamath2021mdetr}. Cap4M and  Cap24M are web-crawled image-text pair dataset collected by GLIP. YFCC1M is our pseudo-labeled image-text pair data sampled from YFCC100M~\cite{thomee2016yfcc100m}. }

\label{tab:dataset_info}
\begin{center}
\resizebox{0.9\linewidth}{!}{
\begin{tabular}{c|cccc}
\toprule
 & Detection & Grounding & Image-text & Total Volume  \\
\midrule
GLIP-T & O365 V1 (0.66M) & GoldG (0.77M) & Cap4M (4M) & \textasciitilde 5.43M\\
GLIP-L & 4ODs (2.66M) & GoldG (0.77M) & Cap24M (24M) & \textasciitilde 27.43 M  \\
\midrule

DetCLIP-T/L & Sampled O365 V2 (0.66M) & GoldG (0.77M) & YFCC1M (1M) & \textasciitilde 2.43M \\

\bottomrule
\end{tabular}
}
\end{center}
\vskip -0.1in
\end{table}

\begin{table}[ht!]
\caption{Manually designed prompts for six downstream detection datasets.}

\label{tab:manual prompts}
\begin{center}
\resizebox{0.8\linewidth}{!}{
    \begin{tabular}{c|cc}
    \toprule
    Dataset & Original Prompt & Manually Designed Prompt    \\
    \midrule
    EgoHands & hand & fist. \tabularnewline
    \midrule
    \multirow{2}{*}{NorthAmericaMushrooms}&  CoW  &  oyster mushroom.\\
    & chanterelle & yellow mushroom.
    \tabularnewline
    \midrule
    Packages & package & a package on the floor. \tabularnewline
    \midrule
    Pothole&  pothole  &  a pit, a sizeable hole. \\
    \midrule
    {Pistols}&  {pistol}  &  pistol or firearm. \\
    \midrule
    \multirow{2}{*}{ThermalDogsAndPeople}&  dog  &  dog in heatmap.\\
    & person & person in heatmap.\\
    \bottomrule
    \end{tabular}
}
\end{center}
\vskip -0.1in
\end{table}



\section{More Results on LVIS and 13 Detection Datasets}

{\textbf{More results under GLIP-protocal.}} We study the generalization ability of our models by zero-shot transferring them to LVIS~\cite{gupta2019lvis} full validation set. 
Following GLIP \cite{li2021grounded}, we use manually designed  prompts for some downstream datasets, as illustrated in Table \ref{tab:manual prompts}.   
AP for LVIS full validation set is reported in Table \ref{tab:sub_LVIS+13dataset}. 
Despite using much less training data, DetCLIP models can dominate their GLIP's~\cite{li2021grounded} counterparts in most cases (except $AP_f$ on LVIS for DetCLIP-L). Notably, compared to GLIP, DetCLIP considerably boosts the performance for rare categories, which is an important indicator reflecting models' generalization ability for the open-world detection task. 
More AP performances for 13 downstream detection datasets can refer to Table \ref{tab:sub_detail_13dataset}.

\begin{table}[h!]
\begin{centering} 
\caption{{Zero-shot transfer performance on LVIS~\cite{gupta2019lvis} full validation dataset. AP$_r$/AP$_c$/AP$_f$ indicate the AP
values for rare, common, frequent categories. `DH' and `F' in GLIP~\cite{li2021grounded} baselines stand for the dynamic head \cite{dai2021dynamic} and cross-modal fusion. 
}}
\label{tab:sub_LVIS+13dataset}
\par\end{centering}
\begin{center}
\begin{small}
\begin{sc}
\resizebox{\textwidth}{!}{
\begin{tabular}{c|cc|ccc}
\toprule 
\multirow{2}{*}{Model} & \multirow{2}{*}{Backbone} & \multirow{2}{*}{Pre-Train Data} & {LVIS val}  \tabularnewline
 &  &  & {AP} ({AP$_r$ / AP$_c$ / AP$_f$})  \tabularnewline
  \midrule
 {GLIP-T(A)\cite{li2021grounded}} & {Swin-T+DH+F} & {O365} & {12.3} ({6.00 / 8.00 / 19.4})  \tabularnewline
 {GLIP-T\cite{li2021grounded}} & {Swin-T+DH+F} & {O365,GoldG,Cap4M} & {17.2} ({10.1 / 12.5 / 25.2}) \tabularnewline
 {GLIP-L\cite{li2021grounded}} & {Swin-L+DH+F} & {4ODs,O365,GoldG,Cap24M} & {26.9} ({17.1 / 23.3 / \textbf{36.4}})  \tabularnewline
  \midrule
{DetCLIP-T(A)} & {Swin-T} & {O365} & {22.1} ({18.4 / 20.1 / 26.0}) \tabularnewline
{DetCLIP-T(B)} & {Swin-T} & {O365, GoldG} & {27.2} ({21.9 / 25.5 / 31.5})  \tabularnewline
\rowcolor{mypink}{DetCLIP-T} & {Swin-T} & {O365, GoldG, YFCC1m} & {28.4} ({25.0 / 27.0 / 31.6})   \tabularnewline
\midrule
\rowcolor{mypink} {DetCLIP-L} & {Swin-L} & {O365, GoldG, YFCC1m} & {\textbf{31.2}} ({\textbf{27.6} / \textbf{29.6 }/ 34.5})\tabularnewline
\bottomrule
\end{tabular}
}
\end{sc}
\end{small}
\end{center}
\end{table}

\begin{table}[h!]
\begin{centering} 
\caption{{Detailed zero-shot transfer AP of DetCLIP on 13 detection datasets~\cite{li2021grounded}.
}}
\label{tab:sub_detail_13dataset}
\par\end{centering}
\begin{center}
\begin{small}
\begin{sc}
\resizebox{\textwidth}{!}{
\begin{tabular}{c|ccccccc}
\toprule 
{Model} & {Thermal} & {Aquarium} & {Rabbits} & Mushrooms & AerialDrone& PascalVOC& Vehicles \tabularnewline
  \midrule
{DetCLIP-T(A)} & {34.1} & {12.4} & {70.3} & {44.1} &{10.8} &{53.5}&{52.7}\tabularnewline
{DetCLIP-T(B)} & {50.8} & {16.6} & {71.8} & {17.6} &{11.6} &{54.2}&{58.4}\tabularnewline
\rowcolor{mypink}{DetCLIP-T} & {51.3} & {18.5} & {75.2} & \textbf{{69.2}} &{10.9} &{55.0}&{55.7} \tabularnewline
\midrule
\rowcolor{mypink} {DetCLIP-L} & {\textbf{{53.8}}} & {\textbf{{25.6}}} & {\textbf{{75.5}}} & {67.0} &{\textbf{{20.4}}} &{\textbf{{56.7}}}&{\textbf{{64.7}}}\tabularnewline
\midrule
{Model} & {EgoHands} & {Raccoon} & {Pothole} & Pistols & Shellfish & Packages & Avg \tabularnewline
  \midrule
{DetCLIP-T(A)} & {6.1} & {36.5} & {3.0} & {30.9} &{17.9} &{67.2}&{33.8}\tabularnewline
{DetCLIP-T(B)} & {33.8} & {51.9} & {15.2} & {29.2} &{21.8} &{71.5}&{38.8}\tabularnewline
\rowcolor{mypink}{DetCLIP-T} & {34.5} & \textbf{{54.0}} & {14.9} & {31.5} &{24.3} &{68.2}&{43.3} \tabularnewline
\midrule
\rowcolor{mypink} {DetCLIP-L} & {\textbf{{40.6}}} & {{52.6}} & {\textbf{{20.3}}} & {{60.7}} &{\textbf{{43.7}}} &{\textbf{{68.8}}}&{\textbf{{50.0}}}\tabularnewline
\bottomrule
\end{tabular}
}
\end{sc}
\end{small}
\end{center}

\end{table}

{\textbf{More results under VILD-protocal.} To make a more comprehensive evaluation of our method, we also perform experiments under the VILD \cite{gu2021open} protocol, i.e., the method is trained on base categories and then evaluated on novel categories using the original LVIS AP metric. We replace the Objects365 part in our training data with LVIS-base, and GoldG and YFCC1M are still included. Including additional data will lead to somehow unfair comparison with VILD but it is necessary since this is the core component in our method to enable zero-shot capability, which differs from VILD that distills knowledge from a pre-trained CLIP model. Note that we implement DetCLIP using the same training/testing setting as in the paper, and do not use techniques such as large-scale jittering and prompt ensemble which is adopted by VILD to boost the performance. The results are shown in the Table \ref{tab:vild}. Our method (27.3 mAP) outperforms VILD (22.5 mAP) by 4.8\% mAP.
}

\begin{table}[ht!]
\caption{{An comparison with VILD model under VILD protocal.}}
\label{tab:vild}
\begin{center}
\begin{small}
\begin{sc}

\begin{tabular}{c|cc}
\toprule 
\multirow{2}{*}{Model} & \multirow{2}{*}{Backbone} & {LVIS val}  \tabularnewline
 &   & {AP} ({AP$_r$ / AP$_c$ / AP$_f$}) \tabularnewline
  \midrule

 {VILD \cite{gu2021open}} & {ResNet50} & {22.5} ({\textbf{16.1} / 20.0 / 28.3})  \tabularnewline

\midrule
 {DetCLIP} & {ResNet50} & {\textbf{27.3}} ({14.9 / \textbf{25.4} / \textbf{34.8}})  \tabularnewline
\bottomrule
\end{tabular}

\end{sc}
\end{small}
\end{center}
\vskip -0.1in
\end{table}

\section{Ablation Studies}

\textbf{Sequential Formulation with Shuffled Grounding Data (Sec 3.1).}
 The sequential formulation~(e.g., GLIP~\cite{li2021grounded}) is not effective for modeling open-world object detection as a visual-language task since it leads to unnecessary interaction between category names in the attention module. To demonstrate the idea, we randomly shuffle the word order in the grounding training data and report the performance comparison in Table \ref{tab:Sequential}. 
It can be seen that randomly shuffling the word order in the grounding data can even bring a slight improvement ~(i.e., +1.4\% on LVIS minival) on zero-shot transfer AP for the downstream detection task, indicating that the noun phrase is more critical for the detection task, compared to the context information.
Therefore, DetCLIP drops the context information and treats each noun phrase as a paralleled text input, which avoids unnecessary attention among class names and achieves better training efficiency.

\begin{table}[h!]
\begin{centering} 
\caption{{ Performance comparison of sequential formulation with different grounding data~(shuffled word order/original caption).  Zero-shot transfer performance on LVIS~\cite{gupta2019lvis} dataset are reported. 
The model is trained on Objects365~\cite{shao2019objects365} and GoldG datasets.
}}
\label{tab:Sequential}
\par\end{centering}
\begin{center}
\begin{small}
\begin{sc}
\resizebox{\textwidth}{!}{
\begin{tabular}{c|c|cccc}
\toprule 
\multirow{2}{*}{Model} & \multirow{2}{*}{Grounding Data } & {LVIS minival} & {LVIS val} \tabularnewline
 &  & {AP} ({AP$_r$ / AP$_c$ / AP$_f$}) & {AP} ({AP$_r$ / AP$_c$ / AP$_f$}) \tabularnewline
  \midrule
\multirow{2}{*}{Sequential concept form} & {Original caption} & 26.0 (18.0 / 22.8 / 30.3) & 18.9 (11.7 / 15.8 / 25.6)  \tabularnewline
 & {Shuffled Word Order} & \textbf{27.4} (\textbf{18.6} / \textbf{23.8} / \textbf{32.3}) &  \textbf{19.9} (\textbf{12.5} / \textbf{16.4} / \textbf{27.0}) \tabularnewline
\bottomrule 
\end{tabular}
}
\end{sc}
\end{small}
\end{center}
\end{table}

\textbf{Important Role of Class Definition.} In DetCLIP, we augment the class names in the detection dataset with their definitions during both training and inference stage, which is termed as \textbf{concept enrichment}. 
To verify that DetCLIP learns knowledge from class definitions, we compare the performances of including/excluding definitions in text input during the inference stage. Table \ref{tab:test_wo_definiton} reports the results. It can be found that adding definitions to class names can significantly improve the zero-shot transfer performance.


\begin{table}[h!]
\begin{centering} 
\caption{{ Effects of concept enrichment during the inference phrase. Text inputs with and without class definition are studied. Zero-shot transfer performance on LVIS~\cite{gupta2019lvis} dataset is reported. The class definition helps detector better recognize objects.
}}
\label{tab:test_wo_definiton}
\par\end{centering}
\begin{center}
\begin{small}
\begin{sc}
\resizebox{0.9\textwidth}{!}{
\begin{tabular}{c|c|cccc}
\toprule 
\multirow{2}{*}{Model} & \multirow{2}{*}{Text Input} & {LVIS minival} & {LVIS val} \tabularnewline
 &  & {AP} ({AP$_r$ / AP$_c$ / AP$_f$}) & {AP} ({AP$_r$ / AP$_c$ / AP$_f$}) \tabularnewline
  \midrule
\multirow{2}{*}{DetCLIP-T(B)} & {class names} & 30.4 (22.4 / 27.1 / 34.8) & 23.2 (14.3 / 20.7 / 29.9)  \tabularnewline
& {class names + def.} &  \textbf{34.4 (26.9 / 33.9 / 36.3)} &  \textbf{27.2 (21.9 / 25.5 / 31.5)}  \tabularnewline
\bottomrule
\end{tabular}
}
\end{sc}
\end{small}
\end{center}
\end{table}

{\textbf{Impact of pre-trained language models in Concept enrichment.}}
{
During training, we use a pre-trained language model to retrieve a definition in our dictionary for concepts without a direct match in WordNet. We conduct experiments to study how the pre-trained language model in the this process affects the final performance.
Three different settings are considered: 1. do not use language model, i.e., directly adopt the category name as the input for the concepts not in WordNet;  2. use a pre-trained FILIP text encoder; and 3. use a pre-trained RoBERTa as in GLIP. The results are shown in the Table \ref{tab:impact of language model}. 
We can observe that: 1) the concept enrichment procedure can bring significant improvements, ~(e.g., +3.6\% on rare categories) even without using a pre-trained language model; 2) using FILIP can further boost the AP performance from 28.3 to 28.8, while using RoBERTa achieves similar performance with no language model is used.
}

\begin{table}[h]
\caption{{Performance comparison of using different pre-trained text encoders in concept enrichment procedure on LVIS minival dataset. The training dataset is Objects365.}}
\label{tab:impact of language model}
\begin{center}
\begin{small}
\begin{sc}
\begin{tabular}{cc|c}
\toprule
Concepts Enrichment& Pre-trained Text Encoder & LVIS minival \\
\midrule
\xmark&/ & 27.8 (22.2/26.8/29.7) \\
\cmark& None & 28.3 (25.8/27.0/29.9) \\
\cmark& RoBERTa-base & 28.2 (24.5/27.3/29.7) \\
\cmark& FILIP text-encoder & \textbf{28.8 (26.0/28.0/30.0)} \\
\bottomrule
\end{tabular}
\end{sc}
\end{small}
\end{center}
\end{table}

\textbf{Other Important Training Techniques.}
Training a vision-language model that works for the open-world detection task is not easy. We highlight two important training techniques we found in our experiments: (1) using a small learning rate for the pre-trained language backbone, since it helps maintain the language model's knowledge learnt in the large-scale pretraining; and (2) removing the regression loss for non-detection data, since it helps alleviate the negative impact caused by inaccurate localization annotation of grounding/image-text pair data. Table \ref{tab:practical tricks} provides the ablation studies of these techniques.

\begin{table}[ht!]
\caption{Important techniques for training DetCLIP. The learning rate of the image encoder is set to 2.8e-4. Models in this table use sequential formulation as in GLIP \cite{li2021grounded}, since these experiments are conducted during our early-stage exploration.}
\label{tab:practical tricks}
\begin{center}
\begin{small}
\begin{sc}
\begin{tabular}{ccc|c}
\toprule
 Pre-train data & LR (Lang. model)  & Reg. Loss & LVIS (MiniVal)  \\
\midrule
O365 & 2.8e-4 &  Det.  & 15.9 (7.00/11.3/21.5)  \\
O365 & 2.8e-5 &  Det.  & \textbf{23.7} (\textbf{16.6}/\textbf{20.5}/\textbf{27.7})  \\
\midrule
O365, GoldG & 2.8e-4 &  Det  & 22.3 (14.5/17.9/27.6)  \\
O365, GoldG & 2.8e-5 &  Det. GoldG. & 22.9 (15.3/21.5/25.6)  \\
O365, GoldG & 2.8e-5 &  Det  & \textbf{26.0} (\textbf{18.0}/\textbf{22.8}/\textbf{30.3})  \\
\bottomrule
\end{tabular}
\end{sc}
\end{small}
\end{center}
\vskip -0.1in
\end{table}

\section{Qualitative Results}

\begin{figure}
    \centering
    \includegraphics[width=\textwidth]{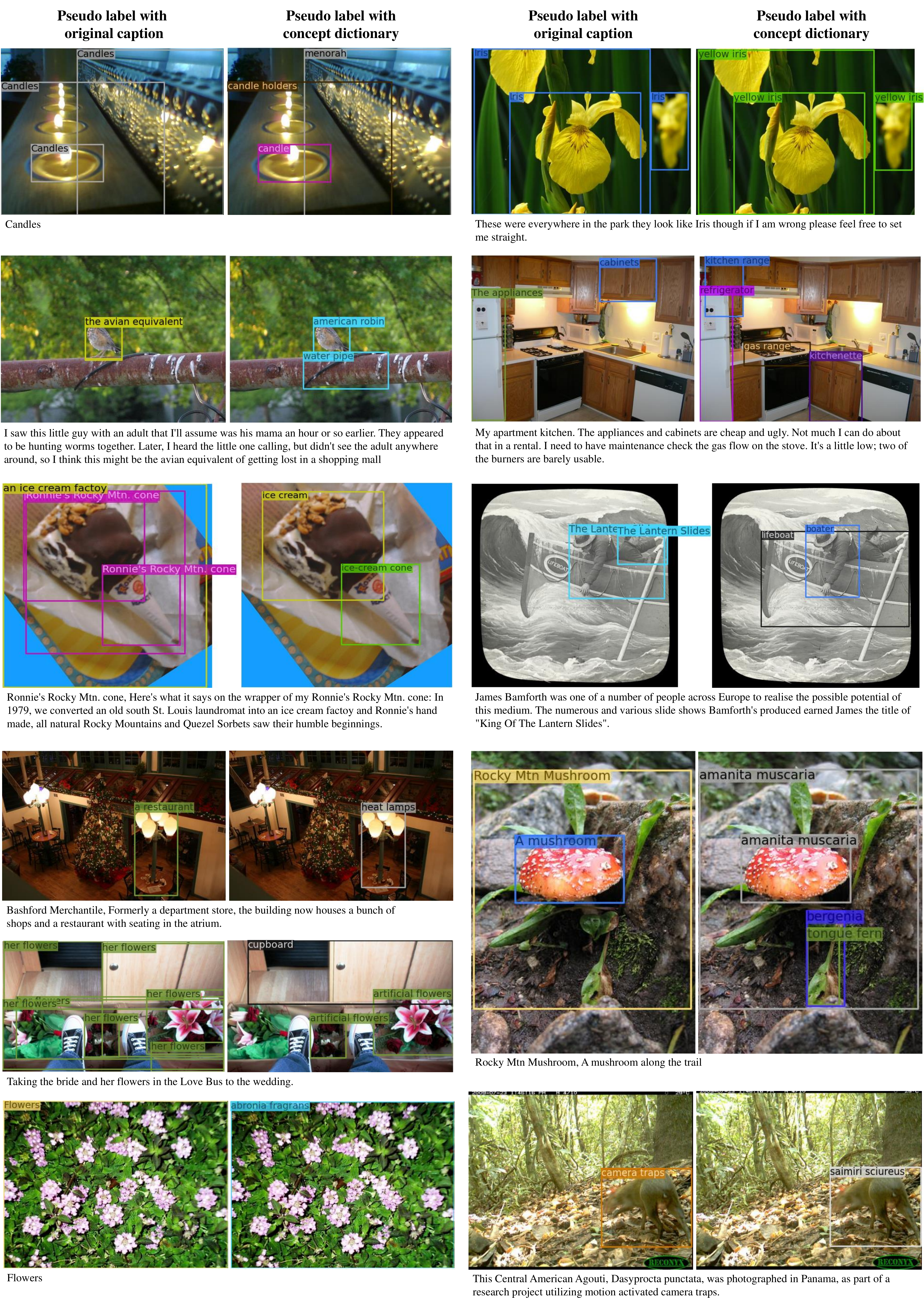}
    \caption{More visualizations of pseudo label results on YFCC~\cite{thomee2016yfcc100m} dataset. The texts below the images are the corresponding captions. Concept dictionary helps produce higher-quality pseudo labels.}
    \label{fig:yfcc_plabel_appendix}
\end{figure}

\textbf{More visualizations of pseudo labels with concept dictionary.} Fig. \ref{fig:yfcc_plabel_appendix} shows extra examples of YFCC data that pseudo labeled with the concept dictionary, as well as their comparisons with the results generated by using the original caption. Concept dictionary alleviates partial-label problem and helps CLIP model provide finer-grained and higher quality pseudo labels. 

\textbf{Retrieval with Concept Dictionary.}
In our \textbf{concept enrichment}, to augment a given class name with its definition, we retrieve it in the constructed concept dictionary. If there is an exact match, we directly use the corresponding definition; otherwise, we use semantic similarity computed by a pre-trained language model to find the closest one. Table \ref{tab:query_def} illustrates some example retrieval results. For class names that are not contained in the dictionary, our method can find proper synonyms.


\begin{table}[ht!]
\caption{ Results of retrieval with the proposed concept dictionary. Class names of Objects365 are used as the queries. Our method can retrieve proper synonyms for class names not contained in the dictionary.  }
\label{tab:query_def}
\begin{center}
\begin{small}
\resizebox{1.0\linewidth}{!}{
\begin{tabular}{cc|l}
\toprule
Query Class Name &  Retrieved Concept & Definition  \\
\midrule

Leather Shoes  &  Boot  &  Footwear that covers the whole foot and lower leg. \\
High Heels  &  Stiletto  &  A woman's shoe with a thin, high tapering heel. \\
Machinery Vehicle  & Truck &  An automotive vehicle suitable for hauling. \\
Cosmetics Mirror  &  Mirror  &  Polished surface that forms images by reflecting light. \\
Induction Cooker  & Hotplate  &  A portable electric appliance for heating or cooking or keeping food warm. \\
Hoverboard  &  Rollerblade  &  Trademark an in line skate. \\

\bottomrule
\end{tabular}
}
\end{small}
\end{center}
\vskip -0.1in
\end{table}


\textbf{Illustrations of Concept Dictionary.}
We illustrate some examples in our concept dictionary in Table \ref{tab:example_concept_dictionary}.
We observe that the concepts collected from the image-text pair data can cover more fine-grained categories~(e.g., cotswold, cuniculus paca) and a wider range of classes~(e.g., giant, cathedral).

\begin{table}[ht!]
\caption{Examples of our concept dictionary. The upper part of the table shows the concepts collected from the detection datasets, while the lower part shows the concepts collected from the image-text pair dataset, i.e., YFCC100M~\cite{thomee2016yfcc100m}.}
\label{tab:example_concept_dictionary}
\begin{center}
\begin{small}
\resizebox{1.0\linewidth}{!}{
\begin{tabular}{c|c}
\toprule
Concept &   Definition  \\
\midrule
\rowcolor{mypink}\multicolumn{2}{c}{Collected from Large-scale Detection Datasets.}\\
\midrule
Cup & A small open container usually used for drinking; usually has a handle.\\
Chips & strips of potato fried in deep fat.\\
Chair & A seat for one person, with a support for the back.\\
Eraser & an implement used to erase something.\\
Gloves & The handwear used by fielders in playing baseball.\\
Recorder & equipment for making records.\\
Street Lights & A lamp supported on a lamppost; for illuminating a street.\\
\midrule
\rowcolor{mypink}\multicolumn{2}{c}{Collected from Image-text Pair Data.}\\
\midrule
Pet & A domesticated animal kept for companionship or amusement.\\
Pod & The vessel that contains the seeds of a plant (not the seeds themselves).\\
Taro & Edible starchy tuberous root of taro plants.\\
Shrub & A low woody perennial plant usually having several major stems.\\
Salmon & Any of various large food and game fishes of northern waters.\\
Brewery & A plant where beer is brewed by fermentation.\\
Pottery & Ceramic ware made from clay and baked in a kiln.\\
Giant & Any creature of exceptional size.giant and any creature of exceptional size.\\
Pagoda & An Asian temple; usually a pyramidal tower with an upward curving roof.\\
Fresco & A mural done with watercolors on wet plaster.\\
Wildflower & Wild or uncultivated flowering plant.\\
Water tower & A large reservoir for water.\\
Basin & A bowl-shaped vessel; usually used for holding food or liquids.\\
Cotswold & Sheep with long wool originating in the Cotswold Hills.\\
Insect & Small air-breathing arthropod.\\
Booth & A table (in a restaurant or bar) surrounded by two high-backed benches.\\
Office & Place of business where professional or clerical duties are performed.\\
Cab & A compartment at the front of a motor vehicle or locomotive where driver sits.\\
Gable & The vertical triangular wall between the sloping ends of gable roof.\\
Hotel & A building where travelers can pay for lodging and meals and other services.\\
Cathedral & Any large and important church.\\
Restaurant & A building where people go to eat.\\
Library & A room where books are kept.\\
Courtyard & An area wholly or partly surrounded by walls or buildings.\\
Footbridge & A bridge designed for pedestrians.\\
Cuniculus paca & Large burrowing rodent of South America and Central America.\\

\bottomrule
\end{tabular}
}
\end{small}
\end{center}
\vskip -0.1in
\end{table}


\newpage

{\small{}\bibliographystyle{plain}
\bibliography{Know_network}

\begin{thebibliography}{10}

\bibitem{brown2020language}
Tom~B Brown, Benjamin Mann, Nick Ryder, Melanie Subbiah, Jared Kaplan, Prafulla
  Dhariwal, Arvind Neelakantan, Pranav Shyam, Girish Sastry, Amanda Askell,
  et~al.
\newblock Language models are few-shot learners.
\newblock In {\em NeurIPS}, 2020.

\bibitem{cai2022x}
Zhaowei Cai, Gukyeong Kwon, Avinash Ravichandran, Erhan Bas, Zhuowen Tu, Rahul
  Bhotika, and Stefano Soatto.
\newblock X-detr: A versatile architecture for instance-wise vision-language
  tasks.
\newblock {\em arXiv preprint arXiv:2204.05626}, 2022.

\bibitem{carion2020end}
Nicolas Carion, Francisco Massa, Gabriel Synnaeve, Nicolas Usunier, Alexander
  Kirillov, and Sergey Zagoruyko.
\newblock End-to-end object detection with transformers.
\newblock In {\em ECCV}, pages 213--229. Springer, 2020.

\bibitem{changpinyo2017predicting}
Soravit Changpinyo, Wei-Lun Chao, and Fei Sha.
\newblock Predicting visual exemplars of unseen classes for zero-shot learning.
\newblock In {\em ICCV}, pages 3476--3485, 2017.

\bibitem{chen2021knowledge}
Jiaoyan Chen, Yuxia Geng, Zhuo Chen, Ian Horrocks, Jeff~Z Pan, and Huajun Chen.
\newblock Knowledge-aware zero-shot learning: Survey and perspective.
\newblock {\em arXiv preprint arXiv:2103.00070}, 2021.

\bibitem{mmdetection}
Kai Chen, Jiaqi Wang, Jiangmiao Pang, Yuhang Cao, Yu~Xiong, Xiaoxiao Li,
  Shuyang Sun, Wansen Feng, Ziwei Liu, Jiarui Xu, Zheng Zhang, Dazhi Cheng,
  Chenchen Zhu, Tianheng Cheng, Qijie Zhao, Buyu Li, Xin Lu, Rui Zhu, Yue Wu,
  Jifeng Dai, Jingdong Wang, Jianping Shi, Wanli Ouyang, Chen~Change Loy, and
  Dahua Lin.
\newblock {MMDetection}: Open mmlab detection toolbox and benchmark.
\newblock {\em arXiv preprint arXiv:1906.07155}, 2019.

\bibitem{dai2016r}
Jifeng Dai, Yi~Li, Kaiming He, and Jian Sun.
\newblock R-fcn: Object detection via region-based fully convolutional
  networks.
\newblock In {\em NeurIPS}, 2016.

\bibitem{dai2021dynamic}
Xiyang Dai, Yinpeng Chen, Bin Xiao, Dongdong Chen, Mengchen Liu, Lu~Yuan, and
  Lei Zhang.
\newblock Dynamic head: Unifying object detection heads with attentions.
\newblock In {\em CVPR}, pages 7373--7382, 2021.

\bibitem{dave2021evaluating}
Achal Dave, Piotr Doll{\'a}r, Deva Ramanan, Alexander Kirillov, and Ross
  Girshick.
\newblock Evaluating large-vocabulary object detectors: The devil is in the
  details.
\newblock {\em arXiv preprint arXiv:2102.01066}, 2021.

\bibitem{devlin2018bert}
Jacob Devlin, Ming-Wei Chang, Kenton Lee, and Kristina Toutanova.
\newblock Bert: Pre-training of deep bidirectional transformers for language
  understanding.
\newblock In {\em Annual Conference of the North American Chapter of the
  Association for Computational Linguistics}, 2019.

\bibitem{dosovitskiy2020image}
Alexey Dosovitskiy, Lucas Beyer, Alexander Kolesnikov, Dirk Weissenborn,
  Xiaohua Zhai, Thomas Unterthiner, Mostafa Dehghani, Matthias Minderer, Georg
  Heigold, Sylvain Gelly, et~al.
\newblock An image is worth 16x16 words: Transformers for image recognition at
  scale.
\newblock In {\em ICLR}, 2020.

\bibitem{du2022learning}
Yu~Du, Fangyun Wei, Zihe Zhang, Miaojing Shi, Yue Gao, and Guoqi Li.
\newblock Learning to prompt for open-vocabulary object detection with
  vision-language model.
\newblock {\em arXiv preprint arXiv:2203.14940}, 2022.

\bibitem{elhoseiny2017link}
Mohamed Elhoseiny, Yizhe Zhu, Han Zhang, and Ahmed Elgammal.
\newblock Link the head to the" beak": Zero shot learning from noisy text
  description at part precision.
\newblock In {\em CVPR}, pages 5640--5649, 2017.

\bibitem{gao2021towards}
Mingfei Gao, Chen Xing, Juan~Carlos Niebles, Junnan Li, Ran Xu, Wenhao Liu, and
  Caiming Xiong.
\newblock Towards open vocabulary object detection without human-provided
  bounding boxes.
\newblock {\em arXiv preprint arXiv:2111.09452}, 2021.

\bibitem{gu2021open}
Xiuye Gu, Tsung-Yi Lin, Weicheng Kuo, and Yin Cui.
\newblock Open-vocabulary object detection via vision and language knowledge
  distillation.
\newblock {\em arXiv preprint arXiv:2104.13921}, 2, 2021.

\bibitem{gupta2019lvis}
Agrim Gupta, Piotr Dollar, and Ross Girshick.
\newblock Lvis: A dataset for large vocabulary instance segmentation.
\newblock In {\em CVPR}, pages 5356--5364, 2019.

\bibitem{he2017mask}
Kaiming He, Georgia Gkioxari, Piotr Doll{\'a}r, and Ross Girshick.
\newblock Mask r-cnn.
\newblock In {\em ICCV}, pages 2961--2969, 2017.

\bibitem{hebart2019things}
Martin~N Hebart, Adam~H Dickter, Alexis Kidder, Wan~Y Kwok, Anna Corriveau,
  Caitlin Van~Wicklin, and Chris~I Baker.
\newblock Things: A database of 1,854 object concepts and more than 26,000
  naturalistic object images.
\newblock {\em PloS one}, 14(10):e0223792, 2019.

\bibitem{jiang2018hybrid}
Chenhan Jiang, Hang Xu, Xiangdan Liang, and Liang Lin.
\newblock Hybrid knowledge routed modules for large-scale object detection.
\newblock In {\em NeurIPS}, 2018.

\bibitem{kamath2021mdetr}
Aishwarya Kamath, Mannat Singh, Yann LeCun, Gabriel Synnaeve, Ishan Misra, and
  Nicolas Carion.
\newblock Mdetr-modulated detection for end-to-end multi-modal understanding.
\newblock In {\em ICCV}, pages 1780--1790, 2021.

\bibitem{kim2021vilt}
Wonjae Kim, Bokyung Son, and Ildoo Kim.
\newblock Vilt: Vision-and-language transformer without convolution or region
  supervision.
\newblock In {\em ICML}, 2021.

\bibitem{kingma2014adam}
Diederik~P Kingma and Jimmy Ba.
\newblock Adam: A method for stochastic optimization.
\newblock {\em arXiv preprint arXiv:1412.6980}, 2014.

\bibitem{krasin2017openimages}
Ivan Krasin, Tom Duerig, Neil Alldrin, Vittorio Ferrari, Sami Abu-El-Haija,
  Alina Kuznetsova, Hassan Rom, Jasper Uijlings, Stefan Popov, Andreas Veit,
  et~al.
\newblock Openimages: A public dataset for large-scale multi-label and
  multi-class image classification.
\newblock {\em Dataset available from https://github. com/openimages}, 2(3):18,
  2017.

\bibitem{Krishna2016}
Ranjay Krishna, Yuke Zhu, Oliver Groth, Justin Johnson, Kenji Hata, Joshua
  Kravitz, Stephanie Chen, Yannis Kalantidis, Li-Jia Li, David~A Shamma,
  Michael Bernstein, and Li~Fei-Fei.
\newblock Visual genome: Connecting language and vision using crowdsourced
  dense image annotations.
\newblock {\em IJCV}, 2016.

\bibitem{krizhevsky2012imagenet}
Alex Krizhevsky, Ilya Sutskever, and Geoffrey~E Hinton.
\newblock Imagenet classification with deep convolutional neural networks.
\newblock In {\em NeurIPS}, volume~25, pages 1097--1105, 2012.

\bibitem{kuznetsova2020open}
Alina Kuznetsova, Hassan Rom, Neil Alldrin, Jasper Uijlings, Ivan Krasin, Jordi
  Pont-Tuset, Shahab Kamali, Stefan Popov, Matteo Malloci, Alexander
  Kolesnikov, et~al.
\newblock The open images dataset v4.
\newblock {\em International Journal of Computer Vision}, 128(7):1956--1981,
  2020.

\bibitem{li2022blip}
Junnan Li, Dongxu Li, Caiming Xiong, and Steven Hoi.
\newblock Blip: Bootstrapping language-image pre-training for unified
  vision-language understanding and generation.
\newblock {\em arXiv preprint arXiv:2201.12086}, 2022.

\bibitem{li2021align}
Junnan Li, Ramprasaath~R. Selvaraju, Akhilesh~Deepak Gotmare, Shafiq Joty, and
  Steven Xiong, Caiming~andHoi.
\newblock Align before fuse: Vision and language representation learning with
  momentum distillation.
\newblock Technical Report arXiv:2107.07651, 2021.

\bibitem{li2019visualbert}
Liunian~Harold Li, Mark Yatskar, Da~Yin, Cho-Jui Hsieh, and Kai-Wei Chang.
\newblock Visualbert: A simple and performant baseline for vision and language.
\newblock {\em arXiv preprint arXiv:1908.03557}, 2019.

\bibitem{li2021grounded}
Liunian~Harold Li, Pengchuan Zhang, Haotian Zhang, Jianwei Yang, Chunyuan Li,
  Yiwu Zhong, Lijuan Wang, Lu~Yuan, Lei Zhang, Jenq-Neng Hwang, et~al.
\newblock Grounded language-image pre-training.
\newblock {\em arXiv preprint arXiv:2112.03857}, 2021.

\bibitem{li2021fully}
Yanwei Li, Hengshuang Zhao, Xiaojuan Qi, Liwei Wang, Zeming Li, Jian Sun, and
  Jiaya Jia.
\newblock Fully convolutional networks for panoptic segmentation.
\newblock In {\em CVPR}, pages 214--223, 2021.

\bibitem{lin2017focal}
Tsung-Yi Lin, Priya Goyal, Ross Girshick, Kaiming He, and Piotr Doll{\'a}r.
\newblock Focal loss for dense object detection.
\newblock In {\em ICCV}, pages 2980--2988, 2017.

\bibitem{Lin2014}
Tsung-Yi Lin, Michael Maire, Serge Belongie, James Hays, Pietro Perona, Deva
  Ramanan, Piotr Doll{\'a}r, and C~Lawrence Zitnick.
\newblock Microsoft coco: Common objects in context.
\newblock In {\em ECCV}, 2014.

\bibitem{liu2016ssd}
Wei Liu, Dragomir Anguelov, Dumitru Erhan, Christian Szegedy, Scott Reed,
  Cheng-Yang Fu, and Alexander~C Berg.
\newblock Ssd: Single shot multibox detector.
\newblock In {\em ECCV}, 2016.

\bibitem{liu2021swin}
Ze~Liu, Yutong Lin, Yue Cao, Han Hu, Yixuan Wei, Zheng Zhang, Stephen Lin, and
  Baining Guo.
\newblock Swin transformer: Hierarchical vision transformer using shifted
  windows.
\newblock In {\em ICCV}, pages 10012--10022, 2021.

\bibitem{Miller1998wordnet}
George~A. Miller.
\newblock Wordnet, an electronic lexical database.
\newblock 1998.

\bibitem{qiao2016less}
Ruizhi Qiao, Lingqiao Liu, Chunhua Shen, and Anton Van Den~Hengel.
\newblock Less is more: zero-shot learning from online textual documents with
  noise suppression.
\newblock In {\em CVPR}, pages 2249--2257, 2016.

\bibitem{radford2021learning}
Alec Radford, Jong~Wook Kim, Chris Hallacy, Aditya Ramesh, Gabriel Goh,
  Sandhini Agarwal, Girish Sastry, Amanda Askell, Pamela Mishkin, Jack Clark,
  et~al.
\newblock Learning transferable visual models from natural language
  supervision.
\newblock In {\em ICML}, pages 8748--8763. PMLR, 2021.

\bibitem{redmon2016you}
Joseph Redmon, Santosh Divvala, Ross Girshick, and Ali Farhadi.
\newblock You only look once: Unified, real-time object detection.
\newblock In {\em CVPR}, 2016.

\bibitem{reed2016learning}
Scott Reed, Zeynep Akata, Honglak Lee, and Bernt Schiele.
\newblock Learning deep representations of fine-grained visual descriptions.
\newblock In {\em CVPR}, 2016.

\bibitem{ren2015faster}
Shaoqing Ren, Kaiming He, Ross Girshick, and Jian Sun.
\newblock Faster r-cnn: Towards real-time object detection with region proposal
  networks.
\newblock In {\em NeurIPS}, 2015.

\bibitem{rezatofighi2019generalized}
Hamid Rezatofighi, Nathan Tsoi, JunYoung Gwak, Amir Sadeghian, Ian Reid, and
  Silvio Savarese.
\newblock Generalized intersection over union: A metric and a loss for bounding
  box regression.
\newblock In {\em CVPR}, pages 658--666, 2019.

\bibitem{salakhutdinov2011learning}
Ruslan Salakhutdinov, Antonio Torralba, and Josh Tenenbaum.
\newblock Learning to share visual appearance for multiclass object detection.
\newblock In {\em CVPR}, 2011.

\bibitem{selvaraju2017grad}
Ramprasaath~R Selvaraju, Michael Cogswell, Abhishek Das, Ramakrishna Vedantam,
  Devi Parikh, and Dhruv Batra.
\newblock Grad-cam: Visual explanations from deep networks via gradient-based
  localization.
\newblock In {\em ICCV}, pages 618--626, 2017.

\bibitem{shao2019objects365}
Shuai Shao, Zeming Li, Tianyuan Zhang, Chao Peng, Gang Yu, Xiangyu Zhang, Jing
  Li, and Jian Sun.
\newblock Objects365: A large-scale, high-quality dataset for object detection.
\newblock In {\em ICCV}, pages 8430--8439, 2019.

\bibitem{socher2013zero}
Richard Socher, Milind Ganjoo, Christopher~D Manning, and Andrew Ng.
\newblock Zero-shot learning through cross-modal transfer.
\newblock {\em NeurIPS}, 26, 2013.

\bibitem{thomee2016yfcc100m}
Bart Thomee, David~A Shamma, Gerald Friedland, Benjamin Elizalde, Karl Ni,
  Douglas Poland, Damian Borth, and Li-Jia Li.
\newblock Yfcc100m: The new data in multimedia research.
\newblock {\em Communications of the ACM}, 59(2):64--73, 2016.

\bibitem{wang2021vlmo}
Wenhui Wang, Hangbo Bao, Li~Dong, and Furu Wei.
\newblock Vlmo: Unified vision-language pre-training with
  mixture-of-modality-experts.
\newblock {\em arXiv preprint arXiv:2111.02358}, 2021.

\bibitem{wang2018zero}
Xiaolong Wang, Yufei Ye, and Abhinav Gupta.
\newblock Zero-shot recognition via semantic embeddings and knowledge graphs.
\newblock In {\em CVPR}, 2018.

\bibitem{xie2021zsd}
Johnathan Xie and Shuai Zheng.
\newblock Zsd-yolo: Zero-shot yolo detection using vision-language
  knowledgedistillation.
\newblock {\em arXiv preprint arXiv:2109.12066}, 2021.

\bibitem{xu2019reasoning}
Hang Xu, ChenHan Jiang, Xiaodan Liang, Liang Lin, and Zhenguo Li.
\newblock Reasoning-rcnn: Unifying adaptive global reasoning into large-scale
  object detection.
\newblock In {\em CVPR}, pages 6419--6428, 2019.

\bibitem{yao2021filip}
Lewei Yao, Runhui Huang, Lu~Hou, Guansong Lu, Minzhe Niu, Hang Xu, Xiaodan
  Liang, Zhenguo Li, Xin Jiang, and Chunjing Xu.
\newblock Filip: Fine-grained interactive language-image pre-training.
\newblock In {\em ICLR}, 2022.

\bibitem{zang2022open}
Yuhang Zang, Wei Li, Kaiyang Zhou, Chen Huang, and Chen~Change Loy.
\newblock Open-vocabulary detr with conditional matching.
\newblock {\em arXiv preprint arXiv:2203.11876}, 2022.

\bibitem{zhang2022glipv2}
Haotian Zhang, Pengchuan Zhang, Xiaowei Hu, Yen-Chun Chen, Liunian~Harold Li,
  Xiyang Dai, Lijuan Wang, Lu~Yuan, Jenq-Neng Hwang, and Jianfeng Gao.
\newblock Glipv2: Unifying localization and vision-language understanding.
\newblock {\em arXiv preprint arXiv:2206.05836}, 2022.

\bibitem{zhang2020bridging}
Shifeng Zhang, Cheng Chi, Yongqiang Yao, Zhen Lei, and Stan~Z Li.
\newblock Bridging the gap between anchor-based and anchor-free detection via
  adaptive training sample selection.
\newblock In {\em CVPR}, pages 9759--9768, 2020.

\bibitem{zhang2021k}
Wenwei Zhang, Jiangmiao Pang, Kai Chen, and Chen~Change Loy.
\newblock K-net: Towards unified image segmentation.
\newblock {\em NeurIPS}, 34, 2021.

\bibitem{zhu2020deformable}
Xizhou Zhu, Weijie Su, Lewei Lu, Bin Li, Xiaogang Wang, and Jifeng Dai.
\newblock Deformable detr: Deformable transformers for end-to-end object
  detection.
\newblock {\em arXiv preprint arXiv:2010.04159}, 2020.

\end{thebibliography}
}{\small\par}

\end{document}